\documentclass{article}
\usepackage{arxiv}

\usepackage{times}
\usepackage{soul}
\usepackage{url}
\usepackage[hidelinks]{hyperref}
\usepackage[utf8]{inputenc}
\usepackage[small]{caption}
\usepackage{graphicx}
\usepackage{amsmath}
\usepackage{amsthm}
\usepackage{booktabs}
\usepackage{float}
\usepackage[ruled,vlined]{algorithm2e}
\usepackage{researchpack}
\usepackage{wrapfig}
\usepackage{natbib}
\usepackage[capitalize,nameinlink]{cleveref} % noabbrev
\usepackage{enumitem}
\usepackage{wrapfig}
\usepackage{xspace}
\usepackage{multirow}
\usepackage{color,soul}
\usepackage{subcaption}
\usepackage{graphicx}
\usepackage{caption}
\usepackage[dvipsnames]{xcolor}

\usepackage{tikz}
\usetikzlibrary{shapes,arrows}

\hypersetup{
    colorlinks=true,
    linkcolor=blue,
    citecolor=blue,
    urlcolor=blue
}

\usepackage{pifont}

\definecolor{plotgreen}{HTML}{00DC00}

\newcommand{\acronym}{Surrogate-based Learning to Guide\xspace}
\newcommand{\method}{\textsc{slog}\xspace}

\newcommand{\datatr}{\ensuremath{\calD_\mathrm{train}}\xspace} % ST: used for training or fine-tuning the original VLM, labeled; used for computing the CE loss
\newcommand{\dataft}{\ensuremath{\calD_\mathrm{tune}}\xspace} % ST: built during the SLOG fine-tuning stage; used for computing the fine-tuning loss

\newcommand{\GuidGen}{\ensuremath{\gamma}\xspace}
\newcommand{\Guidance}{\ensuremath{g}\xspace}
\newcommand{\SurrHuman}{{\sc HumanProxy}\xspace}
\newcommand{\SurrQuality}{\ensuremath{\sigma_\mathrm{quality}}\xspace}
\newcommand{\Score}{\ensuremath{q}\xspace}
\newcommand{\vScore}{\ensuremath{\vq}\xspace}

\newcommand{\CE}{\ensuremath{\mathsf{CE}}\xspace}

\newcommand{\MimicCXR}{\texttt{Mimic-CXR-IV}\xspace}
\newcommand{\CHEXPERT}{\texttt{CheXpert}\xspace}

\newcommand{\RTWOGEN}{\texttt{R2Gen}\xspace}
\newcommand{\RTWOGENCNM}{\texttt{R2GenCMN}\xspace}
\newcommand{\RTWOGENMAMBA}{\texttt{R2GenMamba}\xspace}
\newcommand{\clevr}{{\sc{ClevR}}\xspace}
\title{Learning to Guide Human Decision Makers with Vision-Language Models}

\date{} 					% Or removing it

\author{Debodeep Banerjee\\
	DI, University of Pisa\\
        DISI, University of Trento\\
	\And
	  Stefano Teso  \\
	CIMeC, university of Trento\\
        DISI, University of Trento\\
        \And
	  Burcu Sayin \\
	  DISI, University of Trento \\
	  \And
      Rajib Pal \\
      Bankura Sammilani Medical College\\
      \And
	  Andrea Passerini \\
	  DISI, University of Trento \\
}

\renewcommand{\shorttitle}{Learning to Guide}

\begin{document}
\maketitle
\begin{abstract}
    %There is increasing interest in developing AIs for assisting human decision making in \textit{high-stakes} tasks, such as medical diagnosis, for the purpose of improving decision quality and reducing cognitive strain.
    %
    %Mainstream approaches team up an expert with a machine learning model to which safer decisions are offloaded, thus letting the former focus on cases that demand their attention.
    There is growing interest in AI systems that support human decision-making in \emph{high-stakes} domains (e.g., medical diagnosis) to improve decision quality and reduce cognitive load. Mainstream approaches pair human experts with a machine-learning model, offloading low-risk decisions to the model so that experts can focus on cases that require their judgment.
    This \textbf{\textit{separation of responsibilities}} setup, however, is inadequate for high-stakes scenarios.  The expert may end up over-relying on the machine's decisions due to \textit{anchoring bias}, thus losing the human oversight that is increasingly being required by regulatory agencies to ensure trustworthy AI.  On the other hand, the expert is left entirely unassisted on the (typically hardest) decisions on which the model abstained.
    As a remedy, we introduce \textbf{\textit{learning to guide}} (LTG), an alternative framework in which -- rather than taking control from the human expert -- the machine provides \textit{guidance} useful for decision making, and the human is entirely responsible for coming up with a decision.
    In order to ensure guidance is \textit{interpretable} and \textit{task-specific}, we develop \method, an approach for turning \textit{any} vision-language model into a capable generator of textual guidance by leveraging a modicum of human feedback.
    Our empirical evaluation highlights the promise of \method on both on a synthetic dataset and a challenging, real-world medical diagnosis task.
\end{abstract}

\section{Introduction}

High-stakes applications in healthcare, criminal justice and policy making can substantially benefit from the introduction of AI technology, yet full automation in these scenarios is not desirable, due to ethical, safety and legal concerns, if not explicitly forbidden by law~\citep{Canada2019, AIact2021}.
For these reasons, human-AI or \textbf{\textit{Hybrid Decision Making}} (HDM) is becoming increasingly popular to tackle high-stakes tasks. HDM algorithms pair a human decision maker with an AI agent -- often a machine learning model -- capable of providing support, with the goals of improving \textit{decision quality} and lowering \textit{cognitive effort}.

Most current approaches to HDM follow a principle of \textbf{\textit{separation of responsibilities}}, in the sense that they route novel inputs to exactly one of the two agents -- \textit{either} the human \textit{or} the AI -- who is then responsible for coming up with a decision.
Specifically, in existing approaches \citep{madras2018predict, mozannar2020consistent, keswani2022designing, verma2022calibrated, liu2022incorporating, wilder2021learning, de2020regression, raghu2019algorithmic, okati2021differentiable}, the AI first assesses whether an input can be handled in autonomy -- \eg it is low-risk or can be addressed with confidence -- and defers to a human partner otherwise.
%
%These algorithms are beneficial in that they enable the human to focus on those cases that, according to the machine, most require their attention.
These algorithms help humans focus on the cases the model flags as most needing attention.

\begin{figure*}[!t]
    \begin{center}
        \includegraphics[width=0.85\linewidth]{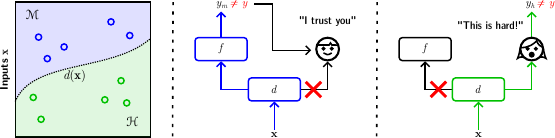}
    \end{center}
    \caption{
    \textbf{Left}:  Existing HDM algorithms employ a deferral function $d(\vx)$ to \textit{partition} the input space $\calX$ into $\calH$ and $\calM$.
    \textbf{Middle}:  A predictor $f(\vx)$ handles the inputs falling in $\calM$ (in \textcolor{blue}{\textbf{blue}}).  Because of \textit{anchoring bias}, the human expert may end up blindly trusting its (possibly poor) decisions $y_m$.
    \textbf{Right}:  The human, on the other hand, is left completely unassisted for those (possibly hard) decisions falling in $\calH$, increasing the chance of mistakes in the human's decisions $y_h$ (in \textcolor{plotgreen}{\textbf{green}}).}
    \label{fig:separation-of-responsibilities}
\end{figure*}

We argue that this setup is \textit{suboptimal} and potentially \textit{unsafe}.
It is suboptimal because, whenever the machine opts for deferral, the human is left resolving hard cases completely unassisted (as in \cref{fig:separation-of-responsibilities}, right).
%This conflicts with the goals of HDM.
%
At the same time, it is unsafe, because humans are affected by \textit{anchoring bias}~\citep{rastogi2022deciding, eigner2024determinants}, a phenomenon whereby decision makers tend to blindly rely on an initial impression (the anchor) and refrain from exploring alternative hypotheses. When the anchor is provided by an algorithm, the bias is amplified as humans tend to over-trust the machine's decisions when available and ignore their own opinions, a phenomenon called  \textit{automation bias}~\citep{cummings2012} (\cref{fig:separation-of-responsibilities}, middle).  This effectively undermines {\em human oversight} over algorithmic decisions, which is increasingly being required by governments around the world to regulate the use of AI in high-stakes applications~\citep{green2022}.

As a remedy, we propose \textbf{\textit{learning to guide}} (LTG), an alternative setup that side-steps these issues. 
In LTG, the machine is trained to supply its human partner with interpretable \textit{guidance} highlighting those aspects of the input that are useful for coming up with a high-quality decision.
For instance, in pathology prediction, the guidance highlights the pathologies present in an input X-ray scan that are indicative of possible diagnoses.
In LTG, \textit{by construction}, all decisions are taken by the human expert -- thus preventing automation bias -- but facilitated by accompanying machine guidance.

% \ST{UPDATE:}
We showcase LTG on \textbf{\textit{medical decision making}} focusing on guidance formulated in \textbf{\textit{natural language}}. Along with that, we also validated the effectiveness of LTG with a synthetic dataset.
To this end, we introduce \method (\acronym), an algorithm for turning large vision-language models (VLMs) \citep{radford2021learning, yan2022clinical, sharma2021medfusenet} into high-quality guidance generators.
In a nutshell, \method takes a VLM pre-trained for caption generation and fine-tunes it using feedback
%in the form of numerical scores representing
about the quality of downstream human decisions inferred from generated guidance.  \method keeps annotation costs under control by training a \textit{surrogate model} that predicts downstream decision quality on a modest amount of feedback, and then using it to fine-tune the VLM in an end-to-end fashion.
Our experiments on a challenging medical diagnosis task indicate that VLMs fine-tuned with \method output interpretable task-specific guidance that can be used to infer high-quality decisions.

\textbf{Contributions.}  In summary, we:
\begin{itemize}[leftmargin=1.25em]
    \item Expose critical limitations in prevailing HDM algorithms that undermine their suitability for high-stakes decision-making.
    %Identify serious flaws with existing HDM algorithms that compromise their applicability to high-stakes tasks.
    \item Propose \emph{learning to guide} (LTG), a novel approach for assisting human decision-makers that keeps humans continuously in the loop.
    %Introduce \textit{learning to guide} (LTG), a novel approach for assisting human decision makers that ensures they are always in the loop.
    \item Present \method, an LTG approach tailored for natural language guidance that can convert large VLMs into interpretable, task-specific guidance generators.
    \item Demonstrate the effectiveness of \method on a challenging medical diagnosis task.
\end{itemize}

\section{Hybrid Decision Making}
\label{sec:hdm}

%We are concerned with decision tasks that, due to safety concerns, cannot be fully automated, such as medical diagnosis.
We target decisions that must retain human oversight (e.g., medical diagnosis) because full automation poses safety risks.\footnote{We focus on classification problems, with inputs $\vx \in \bbR^d$ and categorical or multi-label decisions $y$. Despite this, our remarks apply to other prediction problems as well, \eg regression \citep{de2020regression}.} Research on HDM develops AI assistants to augment human experts on such tasks. Considering the AI assistant and the human expert have different abilities, expertise, and biases, the central question of HDM is how to best integrate them.

%Research on HDM develops expert AI assistants capable of assisting human experts in such tasks.

%

\textbf{HDM by Separation of Responsibilities}.  Existing HDM strategies solve this problem by following a principle of \textit{separation of responsibilities}:  any given instance $\vx$ is assigned to exactly one of the two agents, who is then in charge of decision making, cf.~\cref{fig:separation-of-responsibilities}.
Specifically, they implement a \textit{classifier} $f: \vx \mapsto \hat{y}$, playing the role of an AI agent, as well as a \textit{deferral policy} $d: \bbR^d \to \{{\tt machine}, {\tt human} \}$ that partitions the input domain $\calX$ into two disjoint subsets, $\calM$ and $\calH$.  Novel inputs $\vx$ falling in the former are handled by $f$ and those falling in the latter are handled by the human expert.
This setup is known under a variety of names, including \textit{learning to defer} \citep{madras2018predict,mozannar2020consistent}, \textit{learning under algorithmic triage} \citep{raghu2019algorithmic,okati2021differentiable}, \textit{learning under human assistance} \citep{de2020regression,de2021classification}, and \textit{learning to complement} \citep{wilder2021learning,bansal2021most}.

Approaches differ in how they partition the input space $\calX$.
Earlier methods build on \textit{prediction with a reject option} \citep{cortes2016learning}, in which the deferral policy $d$ observes all incoming instances $\vx$ and offloads those about which the predictor $f$ is unsure (based on, \eg predictive variance) \citep{raghu2019algorithmic}.
Since $f$ is fixed, the partition is static and depends only on the self-assessed uncertainty of the predictor.
Assuming the latter is sufficiently well calibrated \citep{kendall2017uncertainties}, this strategy can perform well in practice \citep{liu2022incorporating}.
The main drawback with this setup is that the partitioning accounts for the machine's performance only, neglecting the human's expertise and biases.
\citet{madras2018predict} improve on this by \textit{learning} the deferral policy $d$ so that it optimizes some decision theoretic measure of \textit{joint team performance}, thus explicitly taking the quality of human decisions into account.
Follow-up works \citep{de2020regression,de2021classification,wilder2021learning} go one step further and train the deferral policy $d$ and the predictor $f$ \textit{jointly}, so as to adapt one to the other. This setup has been extended to incremental \citep{keswani2021towards} and sequential \citep{joshi2021pre} decision making, and to bandit feedback \citep{gao2021human}. Theoretical studies have analyzed the consistency \citep{mozannar2020consistent} and calibration \citep{verma2022calibrated} of the HDM pipeline and the structure of optimal deferral policies \citep{okati2021differentiable}.

\textbf{Issues with Separation of Responsibilities.}  At a high level, an HDM strategy should satisfy the following desiderata:
%the desiderata that an HDM strategy ought to satisfy are the following:
%
\begin{itemize}

    \item[\textbf{D1}.]  \textit{Complementarity}.  It should leverage the complementary capabilities of each agent to obtain better decisions {\em on average}, or equally good decisions at a lower cognitive cost, than each agent individually.

    \item[\textbf{D2}.]  \textit{Synergy}. It should combine the contributions of each agent to obtain better {\em individual} decisions, or equally good decisions at a lower cognitive cost, than each agent individually.

    \item[\textbf{D3}.]  \textit{Reliability}.  It should produce decisions that are more reliable than those made by each agent individually.

\end{itemize}

Existing HDM approaches aim at enabling complementarity (\textbf{D1}).
In fact, the main benefit of offloading decisions to an AI is that of lowering the human's cognitive effort.
Moreover, depending on the relative performance of the predictor on inputs in $\calM$ compared to the expert, they can also improve the quality of the team's decisions (on average across inputs, not necessarily for all inputs).
Under suitable conditions, learning to defer can \textit{provably} do so \citep{donahue2022human}. However, approaches complying with separation of responsibilities completely overlook synergy (\textbf{D2}).
When the machine outputs a decision, the human is tempted to simply stick to it, thus \textit{over-relying on the machine's decisions}, because of the previously mentioned anchoring~\citep{rastogi2022deciding} and automation~\citep{cummings2012} biases.
Conversely, whenever the machine opts for deferral, \textit{the human is left resolving hard cases completely unassisted}.
This also compromises reliability (\textbf{D3}), which is key in high-stakes applications, thus hindering the applicability of HDM.

\begin{figure*}[!t]
    \centering
    \includegraphics[width=0.8\linewidth]{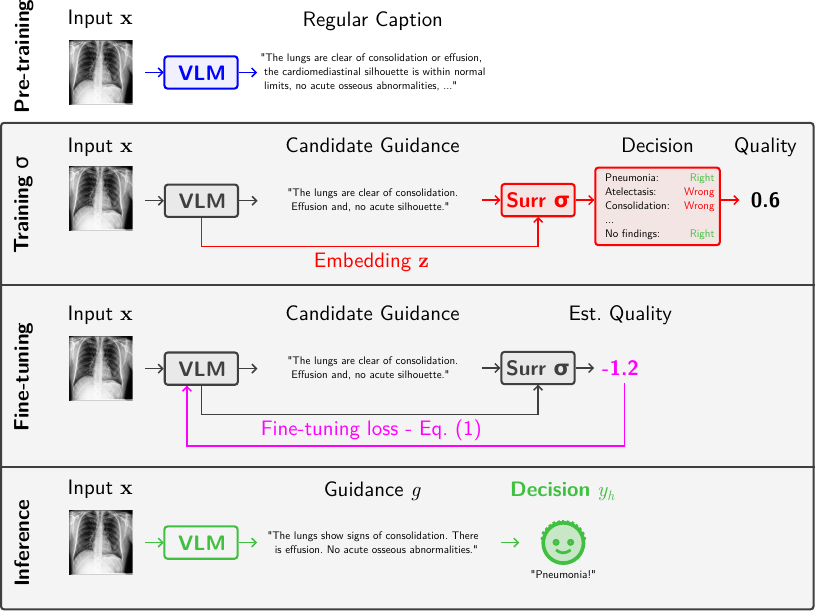}
    \caption{\textbf{The \method approach to learning to guide}.
    \textbf{Tier 1}:  First we take a VLM (in \textcolor{blue}{\textbf{blue}}) pre-trained to generate captions of visual inputs \vx.
    \textbf{Tier 2}:  Next, we train the surrogate \SurrQuality (in \textcolor{red}{\textbf{red}}) to estimate the quality of downstream decisions using a modicum of annotated guidance-quality pairs.  The surrogate takes both images and text (embeddings) as input.
    \textbf{Tier 3}:  Given a trained surrogate $\SurrQuality$, we fine-tune the VLM (in \textcolor{magenta}{\textbf{magenta}}) to output guidance \Guidance achieving high (estimated) decision quality.
    \textbf{Tier 4}:  The fine-tuned VLM (in \textcolor{green}{\textbf{green}}) can readily be used for generating useful textual guidance.
    }
    \label{fig:slog}
\end{figure*}

\section{Beyond Separation of Responsibilities with Learning to Guide}
\label{sec:ltg}

We propose \textit{learning to guide} (LTG), a novel HDM framework that addresses the shortcomings of existing strategies by foregoing separation of responsibilities.
In a nutshell, LTG aims to learn a \textit{guidance generator} $\GuidGen(\vx)$, implemented as a machine learning model, that given an input $\vx$, outputs guidance \Guidance that is useful for \textit{assisting human decision making} on that input.
In medical diagnosis, for instance, given a chest X-ray scan $\vx$, the guidance $\Guidance$ might describe pathology visible in the image that are useful for identifying pathologies and prescribe treatment, as shown in \cref{fig:slog} (top).
%
% \ST{TODO: reference works that support usefulness of decisional guidance.  I CANT FIND GOOD REFERENCES}
%
Critically, and in stark contrast with existing HDM approaches, in LTG the machine does not replace the human:  \textit{the final decision is always taken by the human partner, in collaboration with the AI}.
This means the decision maker is always in the loop and responsible for the final decision.

\textbf{Desiderata for Guidance}.  In order to support human decision making, guidance should satisfy the following natural properties:
\begin{itemize}

    \item[\textbf{D4}.]  \textit{Interpretability}.  It should be \textit{understandable} for the human expert at hand.
 
    \item[\textbf{D5}.]  \textit{Informativeness}.  It should be \textit{informative} for the decision at hand.
    
\end{itemize}
If these are satisfied, then guidance can be used by human experts to address a specific downstream decision making task.
Note that, satisfying these desiderata encourages satisfaction of \textbf{D1}--\textbf{D3}.
In fact, if guidance is interpretable (\textbf{D4}) and extracts decision-relevant elements from the input (\textbf{D5}), it should help the human in taking accurate decisions on individual instances (\textbf{D2}) and as a consequence improving the average quality of the decisions being made (\textbf{D1}).
Additionally, interpretability helps the human in judging the quality of the guidance received, and thus evaluate the reliability of the overall decision (\textbf{D3}).

\textbf{Learning to Guide for VLMs}.  In this paper we focus on \textit{textual guidance} expressed in natural language as a mean to enable interpretability (\textbf{D4}).
Motivated by their state-of-the-art performance in text generation tasks \citep{wei2022emergent} and by their promise in pathological report generation \citep{shamshad2023transformers, chen2020generating, chen2021cross, hou2021ratchet, kayser2022explaining, yunxiang2023chatdoctor, Bazi2023, Drozdov2020, yan2022clinical}, we propose to leverage \textit{vision-language models} (VLM) to generate guidance.

Off-the-shelf VLMs are not conceived for generating guidance for \textit{specific} decision making tasks, and thus violate informativeness (\textbf{D5}).
Clearly, a perfectly accurate medical report is also an optimal guidance for follow-up decisions, but generating highly accurate reports requires massive amounts of supervision, and reports generated by specialized VLMs are far from perfect \citep{shamshad2023transformers}.

The question is then how to \textit{convert} such models into high-quality guidance generators.
Focusing on (medical) decision making from image data, we address this problem by introducing \method, a novel approach for \textit{turning vision-language models into guidance generators using human feedback} designed to comply with \textbf{D1}--\textbf{D5}.
The rationale behind \method is to encourage VLMs to focus on accurately reporting those aspects of the input image that are most relevant \textit{for the follow-up decisions}, possibly overlooking less important details.
Next, we briefly discuss how \method uses annotations and then proceed to outline the main algorithm.

\subsection{Estimating Downstream Decision Quality}
\label{sec:method-surrogate}

% \ST{TODO: stress D1--D5}
Optimizing guidance for synergy (\textbf{D2}) requires knowing the quality of downstream decisions taken by a human expert supplied with the guidance itself.
\method assumes access to quality ratings $\vScore \in [0, 1]^d$, where each $\Score_i$ encodes the quality of a downstream decision.  For instance, if the expert has to determine the state of two conditions (e.g., ``{\tt pneumonia}'' and ``{\tt fracture}''), then $d = 2$.
Quality ratings for expert decisions can be obtained by comparing these against a gold standard (using, e.g., decision accuracy) or by consulting a second expert (using, e.g., a star rating system).

Clearly, there is a tension between the number of annotations necessary for fine-tuning a VLM and the cost of eliciting such annotations.
\method addresses this issue by training a \textit{surrogate model} $\SurrQuality: (\vx, \vz) \mapsto \widehat{\vScore}$ using a modicum of annotated quality ratings, and using it to estimate the quality of guidance $\Guidance$ generated by the VLM during fine-tuning.
In practice, \method fits the surrogate on a training set $\calD_{\mathrm{surr}} = \{ (\vx_i, \vz_i, \Score_i) \}$, where $\vx_i$ is an input image, $\vz_i$ is the embedding of the VLM's guidance $\Guidance_i$ for that input, and $\Score_i$ is the quality of that guidance, by minimizing an average cross-entropy loss of the form:
\[
    % \textstyle
    \frac{1}{|\calD_{\mathrm{surr}}|} \sum_{( \vx, \vz, \vScore ) \in \calD_{\mathrm{surr}}} \ \frac{1}{d} \sum_{i=1}^d \CE( \Score_i, \SurrQuality(\vx, \vz)_i )
\]

\subsection{The SLOG Loop}

In essence, \method takes a pre-trained VLM caption generator \GuidGen and fine-tunes it for a number of rounds $T$.
Let $\datatr$ be a data set of image-caption pairs (for instance, a subset of the data that \GuidGen was trained on) and $\dataft$ a larger set of \textit{unlabeled} images from the target decision making task.
In each round $t = 1, \ldots, T$, \method samples a batch $\{ \vx_1, \ldots, \vx_B \}$ from $\dataft$ uniformly at random with replacement, and computes guidance $\Guidance_i^t = \GuidGen(\vx_i)$ and embeddings $\vz_i^t$ for each input.
Then, it evaluates the quality of the generated guidance using the frozen surrogate \SurrQuality and fine-tunes the VLM \GuidGen by minimizing an augmented loss of the form: 
\[
    % \textstyle
    \CE(\gamma, \datatr) -
        \frac{\lambda}{|\dataft|} \sum_{( \vx, \vz) \in \dataft} \frac{1}{d} \sum_{i=1}^d \SurrQuality(\vx, \vz)_i
    \label{eq:fine-tune-llm}
\]
% \DB{We don't have any fine-tuned split. Proposed eqn:\[
%     \textstyle
%     \CE(\gamma, \datatr) -
%         \frac{\lambda}{|\datatr|} \sum_{( \vx, \vz) \in \datatr} \frac{1}{d} \sum_{i=1}^d \SurrQuality(\vx, \vz)_i\]}
%
for a given number of epochs.
\cref{eq:fine-tune-llm} trades off text generation performance on the training set $\datatr$ -- so as to discourage catastrophic forgetting -- and estimated quality of downstream decisions on the fine-tuning set $\dataft$.  Here, $\lambda > 0$ is a hyper-parameter.
Fine-tuning then amounts to applying gradient descent to batches comprising training and fine-tuning examples in equal proportions.
Once done, the \method loop repeats.
As long as the surrogate generalizes the quality rating annotations, the VLM gradually learns to output text that works well as guidance tailored for the target decision task.
%
%\ST{update based on experiments:} Since the VLM's embedding space changes during fine-tuning, \method keeps the surrogate up to date by retraining it after each iteration.  This operation is very efficient compared to fine-tuning the VLM itself.

\subsection{Benefits and Limitations}
\label{sec:benefits-limitations}

In stark contrast with existing HDM strategies, \method ensures that the human receives guidance useful for decision making while keeping them in the loop.
The cognitive load of LTG is entirely devoted to ensuring it can be safely employed in high-stakes applications, where there is little room for mistakes and humans \textit{have} to be in control at all times \citep{zhang2020effect}, as increasingly prescribed by legal frameworks \citep{Canada2019, AIact2021}.  LTG and \method are designed explicitly for supporting HDM in these cases.
\method is reminiscent of mainstream approaches to LLM alignment, such as reinforcement learning with human feedback (RLHF) \citep{ziegler2020finetuning, ouyang2022training}, but differs from them in aims and technology.  While RLHF strives to improve factuality and reduce harmfulness of generated content \citep{ouyang2022training}, ignoring the decisions these impinge on, \method specifically aims at improving quality of downstream human decisions for a specific decision making task.  At the same time, \method foregoes reinforcement learning approaches \citep{schulman2017proximal} in favor of a simpler and more direct end-to-end fine-tuning strategy.

One limitation of \method is that the performance of the guidance generator hinges on that of the surrogate, which in turn relies on the amount of quality ratings available for training.  In \cref{sec:experiments} we present an ablation study showing that a limited amount of quality annotations are sufficient for \method to improve generated guidance.
Another limitation of \method is that it currently assumes quality ratings are readily available, which is not always the case.  One option is then to integrate \method with active learning strategies \citep{settles2012active, herde2021survey} to acquire informative quality ratings whenever needed.  Doing so is however outside the scope of this paper and left to future work.
Finally, the guidance output by VLMs may suffer from hallucination, that is, it may contain untrue statements.  However, \method directly maximizes factuality of guidance on the fine-tuning set \dataft, meaning that a simple way of reducing the chance of non-factual statements is to employ a larger fine-tuning set.  This is relatively cheap to do, as no annotations are required. Moreover, large language models can be surprisingly well-calibrated \citep{kadavath2022language}, meaning that generated guidance can be filtered based on the VLM's own uncertainty estimates to prevent over-reliance \citep{eigner2024determinants} and avoid low-quality decisions \citep{zhang2020effect}.

\section{Empirical Analysis}
\label{sec:experiments}

In this section, we investigate the following research questions:
\begin{itemize}[leftmargin=2em]
    \item[\textbf{Q1}] Does \method work in a controlled setting? %\ST{not a good research question}
    \item[\textbf{Q2}] Does \method improve generated guidance despite relying on a surrogate model?
    \item[\textbf{Q3}] Does \method improve the quality of the decisions made using its guidance?
    \item[\textbf{Q4}] Does \method help improve performance even when examined by a human physician?
\end{itemize}
In order to answer \textbf{Q1}, we investigate the effectiveness of \method in a controlled experimental setup using the ClevR dataset \citep{johnson2017clevr}.
For the remaining questions, we employ a real world medical decision making dataset.
We discuss the specifics of all experiments in the following.
We will make all source code public upon acceptance of the manuscript.

% \begin{figure}[!t]
%     \centering
%     \begin{tabular}{ccc}
%         \includegraphics[width=0.45\linewidth]{./figures/clevr_example.pdf}
%         &&
%         \includegraphics[width=0.45\linewidth]{./figures/example_new.pdf}
%     \end{tabular}
%     \caption{\textbf{Left}: Example \clevr image and corresponding textual description \AP{@debodeep: make the image comparable in size and font to the right one}. \textbf{Right}: Example of \MimicCXR radiograph and corresponding medical report, comprising \emph{findings} and\emph{impression} sections.}
%     \label{fig:sidebyside}
% \end{figure}

\begin{figure}
    \centering
    \includegraphics[scale=0.4]{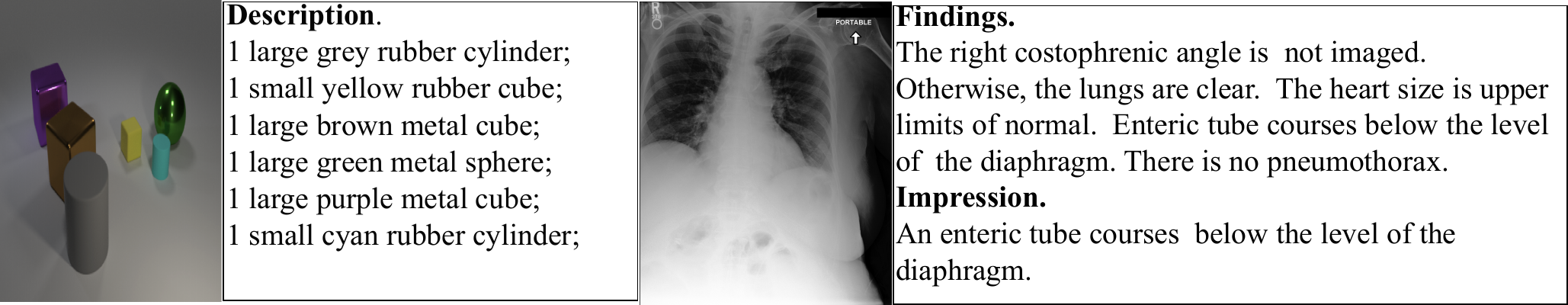}
    \caption{\textbf{Left}: Example \clevr image and corresponding textual description. \textbf{Right}: Example of \MimicCXR radiograph and corresponding medical report, comprising \emph{findings} and\emph{impression} sections.}
    \label{fig:sidebyside}
\end{figure}

\subsection{Q1: The ClevR Task}
\label{sec:experiments-clevr}

\textbf{Data set}.  We first evaluate \method on a variant of the \underline{\clevr} \citep{johnson2017clevr} dataset.  This consists of $60k$ images for training, $10k$ for validation and $15k$ for testing. Each image represents several three-dimensional objects with different shapes, colors, sizes, materials, and positions, and comes with a structured (non-textual) description of its contents.  We translate these into natural language to obtain ground-truth textual descriptions, see \cref{fig:sidebyside} (left) for an example.

\textbf{Decision-making task}.  We are interested in evaluating \method's ability of producing high-quality guidance for decision-makers. Since the \clevr data set comes with no pre-specified decision task, we define one ourselves. Specifically, we construct the following six rules about the presence of certain objects in the image:
\begin{description}[leftmargin=4em]
    \item[Rule 1] Does the image contain one large green sphere \textbf{or} one small rubber cube?
    \item[Rule 2] Does the image contain one large red sphere \textbf{or} one green object?
    \item[Rule 3] Does the image contain one large rubber cube \textbf{and} one sphere?
    \item[Rule 4] Does the image contain one rubber cylinder \textbf{and} two small objects?
    \item[Rule 5] Does the image contain one small red metal cube \textbf{or} two rubber cylinders?
    \item[Rule 6] Does the image contain one sphere \textbf{and} two small metal objects?
\end{description}
Each rule allows us to define positive images (\ie those that satisfy the rule) and negative images (\ie those that do not), yielding six labels per image.
We construct these rules based on the frequency of object features in the images, selecting conditions with a balanced positive/negative ratio.
Given an image, the decision task amounts to predicting what rules will fire.

%In an ideal scenario, as \method's job is to guide the human in the down-stream decison making task, the human should read the guidance provided with \method and take the decisions, \ie identify the availability of the binary rules.
In a nutshell, in this experiment we first train a VLM on the ground-truth (decision-agnostic) captions, then construct a surrogate model that simulates the human decision-maker, and then fine-tune the VLM with \method (\Cref{eq:fine-tune-llm}) to produce guidance.  Next, we detail each part of the pipeline.

\textbf{Vision-language Model}.  We apply \method to a vision encoder-decoder model with a simple transformer \citep{vaswani2017attention} model.  Given an image $\vx$, the VLM outputs a textual caption describing what objects it contains.  We trained the model with an nVidia A100 40 GB GPU with a batch size of 256, restricting the maximum length to 55. {With a patience of 25, we trained the model for 30 epochs and retained the model achieving the best BLEU$_4$ score.}
% We kept a patience of 25 and retained the model achieving the best BLEU$_4$ score. 
The BLEU$_k$ score is useful to evaluate the quality of machine-generated text with respect to a reference text, usually generated by a human. Specifically, BLEU$_k$ considers the overlap of each $k$-gram between the machine-generated and the reference text \citep{papineni2002bleu}. 

% Assuming the availability of ground truth information for a subset of data (in real life scenario), we train \SurrHuman on $10\%$ of training data only. Note that, both the \SurrHuman and the \SurrQuality are trained on the same 
% In this experiment, \method aims at generating higher-quality guidance that helps human decision makers to correctly classify images belonging to different classes.

\textbf{Simulating the human expert}. To retain full control over the experimental setting, we \textit{simulate} the human expert using a machine learning classifier, denoted \SurrHuman, which takes as input an image $\vx$ and the corresponding guidance $\Guidance$ and uses them to infer the six binary labels.
The surrogate \SurrHuman is a multi-modal model with a \textsc{ResNet101} model as visual extractor \citep{he2016deep} and a transformer module as textual encoder.
The sole purpose of this surrogate is to produce (possibly imperfect) decisions for all six rules for the entire data set, which are not provided by the original \clevr dataset but are needed for training the \method surrogate model \SurrQuality.

Specifically, we obtain the prediction using a $k$-fold cross validation procedure: in each fold, we fit \SurrHuman using the ground-truth labels (rule activations) from the training split, and produce predictions for all examples in the test split.  Then we aggregate all test prediction to annotate the whole datasets.  This step ensures there is no data leakage between train and test splits. {In our experiental set-up, we considered $k=5$.}

%Moreover, we train \SurrHuman on $10\%$ of training split only, to ensure it achieves reasonable but sub-optimal performance, to mimic an imperfect decision maker. \ST{POINT OF CONCERN: but we evaluate the performance of \method using a perfect decision maker (the NLP pipeline described below).} \DB{in the mimci case, \method is evaluated in somewhat similar matter where chexpert being a rule based classifier (still predictive though). However, as in ClevR we don't have anything as such, we had to reply on an NLP tool. Nevertheless, is it really a concern as \method, along with other competitors are being evaluated with the same tool.} \ST{I'm not suggesting we should change the experiments; I'm suggesting we should find a nice way of saying it :-)}

% After training \SurrHuman on the ground-truth labels, we compare its predictions against the ground-truth labels themselves to obtain the per-example correctness annotations necessary to train \SurrQuality. 

% We used $k$-fold cross validation and compared the predictions and ground-truth annotations in the validation dataset in order to produce the training data of the \SurrQuality. .\ST{@DB: clarify performance and architecture a bit.} 

\textbf{Quality surrogate model}. The surrogate \SurrQuality takes the same inputs as \SurrHuman, but estimates the quality of the predictions made by \SurrHuman.
To this end, we compare said predictions with the ground-truth rule activations and record whether they match or not.  Then we train \SurrQuality to predict the \textit{correctness} of predictions given only the image $\vx$ and guidance $\Guidance$.  We employed $70\%$-$20\%$-$10\%$ splits for training, validation and test, respectively. We trained \SurrQuality with a cross entropy loss to predict the accuracy of the predictions, and selected the model that attained the best micro validation $F_1$. We set patience to $20$ epochs, for a total of $75$ training epochs.

\textbf{Applying {\sc \method}}.  Given the trained surrogate \SurrQuality, we ran the \method finetuning for 10 epochs. In this phase, we froze both \SurrQuality and the visual encoder of the VLM. We evaluated several values of the hyperparameter $\lambda$ (cf. \cref{eq:fine-tune-llm}) and chose $\lambda = 1$ as it yielded the highest $F_1$ score on a validation split.

\textbf{Competitors and metrics}.  We assess the performance of \method both quantitatively and qualitatively, and compare it against two competitors.
These include the original pretrained VLM (denoted ``baseline'') as well as the same VLM fine-tuned for the same number of epochs as the \method variant, but with $\lambda = 0$, so as to disable the \method loss term (denoted ``fine-tuned'').

We evaluate both the quality of down-stream decisions taken using the generated guidance and the textual guidance itself.
For the former, we report the test set precision (Pr), recall (Rc), and $F_1$ score of the decisions, both per-label and (micro- and macro-) averaged over all six labels.  We obtain the down-stream decision by applying the rules to the generated guidance, using a simple NLP pipeline that scans textual guidance for presence of individual objects and their properties (\eg ``a large red sphere'') and checks which rules fire based on the patterns it matched.  We compare the resulting decisions against the ground-truth labels.
For the latter, we report the $BLEU_k$ score for $k = 1, \ldots, 4$ with respect to the ground-truth captions, the average guidance length, and the average estimated quality output by \SurrQuality as a sanity check.

\begin{table}
    \caption{\textbf{{\sc \method} boosts estimated quality of generated guidance without compromising text quality in \clevr}. The results show that \method substantially improves estimated guidance quality as measured by the surrogate model ($\SurrQuality$) without affecting text quality as measured by BLEU scores over ground-truth caption data.}
    \label{tab:slog_clevr_lang}
    \centering
    \begin{tabular}{lcccccc}
        \toprule
        {\sc Model}
        & {\sc Bleu$_1$}
        & {\sc Bleu$_2$}
        & {\sc Bleu$_3$}
        & {\sc Bleu$_4$}
        & {\sc BleuRT}
        & $\SurrQuality$
        \\
        \midrule
        Baseline
        & 0.95 & 0.92 & 0.89 & 0.85 & 0.73 & 1.96 \\
        Fine-tuned
        & 0.94 & 0.92 & 0.89 & 0.85 & 0.72 & 1.95 \\
        \method
        & \textbf{0.98} & \textbf{0.96} & \textbf{0.93} & \textbf{0.88} & \textbf{0.76} & \textbf{2.28} \\
\bottomrule
\end{tabular}
\end{table}

\begin{table}
	\centering
	\caption{\textbf{{\sf \method} boosts the quality on the downstream decision in \clevr}, as shown by the precision, recall and $F_1$ performance of decisions entailed by \method's guidance compared to that of decisions entailed by the baseline VLM and a VLM fine tuned without the \method loss.  Best $F_1$ results in \textbf{bold}.}
	\label{tab:clevr-rules-performance}
	\begin{tabular}{lccccccccc}
		\toprule
		& \multicolumn{3}{c}{Baseline} & \multicolumn{3}{c}{Baseline (Fine-tuned)} & \multicolumn{3}{c}{\method} \\
		\cmidrule(lr){2-4} \cmidrule(lr){5-7} \cmidrule(lr){8-10}
		\textbf{Rule} & Pr & Rc & $F_1$ & Pr & Rc & $F_1$ & Pr & Rc & $F_1$ \\
		\midrule
		Rule 1 & 92.33 & 95.91 & 94.09 & 93.55 & 94.03 & 93.79 & 94.52 & 97.71 & \textbf{96.09} \\
		Rule 2 & 99.98 & 97.47 & 98.71 & 100.00 & 95.84 & 97.88 & 99.92 & 99.05 & \textbf{99.49} \\
		Rule 3 & 99.75 & 84.93 & 91.75 & 99.75 & 84.89 & 91.72 & 99.77 & 90.09 & \textbf{94.68} \\
		Rule 4 & 94.44 & 96.67 & 95.54 & 96.74 & 94.63 & 95.67 & 96.97 & 98.33 & \textbf{97.65} \\
		Rule 5 & 95.47 & 93.38 & 94.41 & 98.05 & 92.25 & 95.06 & 97.91 & 96.25 & \textbf{97.07} \\
		Rule 6 & 86.59 & 98.26 & 92.05 & 93.88 & 94.70 & 94.28 & 93.36 & 99.45 & \textbf{96.31} \\
		\midrule
		\textsc{Macro Avg} & 94.76 & 94.44 & 94.42 & 97.00 & 92.72 & 94.73 & 97.07 & 96.82 & \textbf{96.88} \\
		\textsc{Micro Avg} & 94.42 & 95.14 & 94.78 & 96.84 & 93.32 & 95.05 & 96.94 & 97.31 & \textbf{97.12} \\
		\bottomrule
	\end{tabular}
\end{table}

\textbf{Q1: {\sc \method} improves the quality of downstream decisions in our controlled setting}.  Our results are summarized in \Cref{tab:slog_clevr_lang,tab:clevr-rules-performance}.
\Cref{tab:slog_clevr_lang} shows that \method's guidance yields a distinct improvement in terms of $\SurrQuality$ score (second to last column), as expected.  This highlights how the \method fine-tuning procedure succeeds in optimizing the \method loss (\cref{eq:fine-tune-llm}).
The remaining results suggest that doing so yields better guidance.
In fact, \method also out-performs both competitors in terms of textual quality: it achieves between $+3$ and $+4$ \% in terms of BLEU scores and the BLEURT\footnote{While the BLEU scores measure the $k$-gram based overlap between the predicted and generated texts, the BLEURT is a BERT \citep{devlin2019bert} based regression model trained on human-ratings data.} \citep{sellam2020bleurt} metric (first five columns) compared to both the pretrained model and its fine-tuned variant.   

Simultaneously, \Cref{tab:clevr-rules-performance} indicates that the guidance produced with \method facilitates inferring correct labels compared to the captions produced by the baseline models.  This holds for each rule/label individually (first six rows), and on average (last two rows).  This provides initial evidence that \method also improves the quality of down-stream decisions in a controlled setting.

\begin{figure*}
    \centering
    \includegraphics[width=1.0\linewidth]{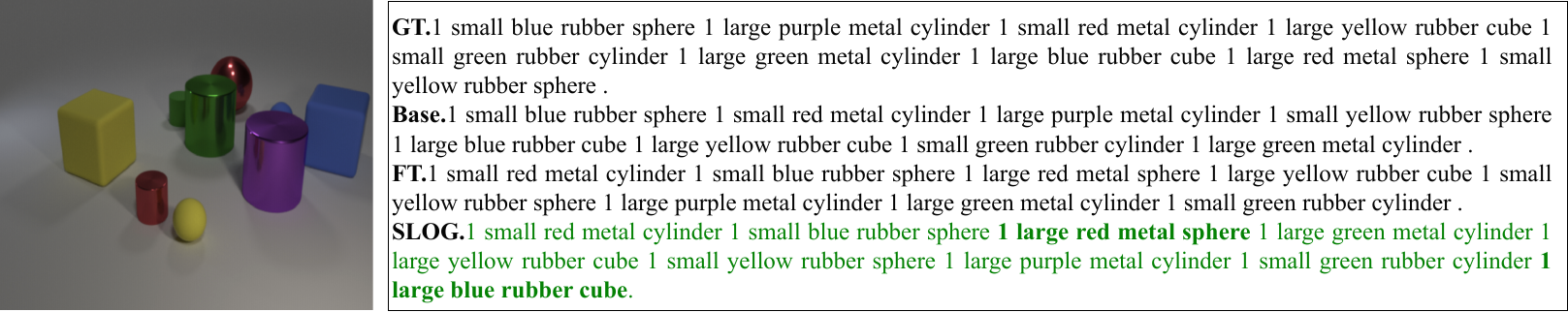}
    \caption{\textbf{Example guidance generated by different competitors on \clevr.} While \method neither misses nor hallucinates any description, each of the other two competitors, despite not hallucinating, misses one description. The missed descriptions are presented in green bold fonts in the above text.}
    \label{fig:clevr_output}
\end{figure*}

To further illustrate the benefits of \method, we report in \Cref{fig:clevr_output} an example of the guidance it produces, compared to the captions output by the competitors.  The example shows how \method's guidance describes objects that are relevant for the decision correctly that the competitors neglect.

\begin{table}
    \caption{Comparison between the original and our filtered splits for the \MimicCXR dataset.}
    \label{tab:mimic_data}
    \centering
    % \scriptsize
    \begin{tabular}{lcccccc}
        \toprule
        & \multicolumn{3}{c}{\sc Original split} & \multicolumn{3}{c}{\sc  Our split}\\
        \cmidrule(lr){2-4}  \cmidrule(lr){5-7}
	\sc  Components & \sc  Train & \sc  Val & \sc  Test & \sc  Train & \sc Val & \sc  Test\\
        \midrule		
        \sc Reports &\sc 222,758&\sc 1,808& \sc 3,269 &\sc 125,417&991&1,624 \\
        \sc Images  &\sc 368,960 & \sc 2,991 & \sc 5,159 &232,855&1,837&2,872 \\
	\bottomrule
    \end{tabular}
\end{table}

\subsection{Q2--Q4: The \MimicCXR Task}

\textbf{Data set}.  Next, we evaluate \method on the \underline{\MimicCXR} data set \citep{johnson2019mimic}, one of the largest publicly available medical decision data sets, consisting of $227,835$ radiology reports and $377,110$ chest X-ray scans. We focus on the \textit{findings} and the \textit{impression} sections of the reports.  As shown in \cref{fig:sidebyside} (right) the findings are text-based descriptions of what can be observed in the scan, and constitute the basis on top of which the expert forms their impression, \ie their initial opinion about the potential pathologies of a patient.
%and constitute the guidance that is expected from the machine,
%while the impressions are summaries of the findings focusing on aspects which are most relevant for medical decision-making, and closely match what an expert would extract from the
%guidance provided by the
%findings in order to make decisions.  
We discarded examples where either findings or impressions were not available, resulting in the \textit{training}, \textit{validation} and \textit{test} splits presented in \cref{tab:mimic_data}. %\DB{worth mentioning? we adopted \MimicCXR's official train, test and validation split}
%
%We use the full training split for fine-tuning the VLM, but \ST{stress more} only \ST{@DB} $10\%$ of it (the \textit{surrogate} split) for fitting our surrogate models.

\textbf{Vision-language Models}.  We apply \method to {three} vision-language models: \RTWOGEN \citep{chen2020generating}, \RTWOGENCNM \citep{chen2022cross} {and, \RTWOGENMAMBA \citep{sun2025r2gen}.}
%
% The former is a memory-driven transformer \citep{vaswani2017attention} specifically designed for pathological report generation from chest X-ray images, while the latter uses a cross modal network (CMN) in order to achieve better mapping between diverse modalities. 
{\RTWOGEN is a memory-driven transformer \citep{vaswani2017attention} specifically designed for pathological report generation from chest X-ray images, while \RTWOGENCNM uses a cross modal network (CMN) in order to achieve better mapping between diverse modalities. Finally, \RTWOGENMAMBA leverages a recent \textsc{Mamba} state-space model in its encoder architecture.}
The \RTWOGEN architecture builds on the observation that similar radiographs may correspond to reports sharing similar patterns.  To exploit this, it employs a pre-trained CNN model to extract patch features and encodes these into hidden states with an encoder.  A decoder then maps the hidden states into words at each time point with the help of a relational memory and memory-driven conditional layer normalization.  The relational memory component allows the transformer to store and repurpose shared patterns and thus generate more coherent reports \citep{chen2020generating}.
\citet{chen2022cross} argues that the existing literature offers only limited scope for proper alignment across modalities. Addressing this issue, the authors developed \RTWOGENCNM, a cross modal network where the encoded features of an image is fed to the CMN module to obtain the memory representations. A similar operation is done for the text embeddings. Thus, the shared information of the text and visual features can be stored in the memory. In particular, the CMN module employs a matrix where each row of the matrix is allotted for cross-modal memory information for image and texts.

% \underline{\RTWOGENCNM} \citep{chen2021cross} extends \RTWOGEN with an additional memory matrix for cross modal information storage.  This is used to perform memory querying and memory responding. At the time of query, the most relevant memory vectors are extracted and based on the visual and textual features, their corresponding weights are computed. Based on the weights of the queries vectors, responses are created and eventually the responses are passed to the the encoder-decoder layer.

%\AP{I commented out the implementation details, they don't seem useful}
% We trained our baseline VLM with an Nvidia A40 GPU and with batch size 64 and fine-tuned the baseline model with an Nvidia A100 GPU with a batch size of 32. Our choice of hyper-parameters remained same as used by \citep{chen2020generating}. While finetuning, we optimized the loss mentioned in \ref{eq:fine-tune-llm}. We run the finetuning for additional 10 epochs and used the model on the last epoch. We tried multiple values of the hyper-parameter $\lambda$ and chose it to be 10 which resulted in the best outcome. Along with fine-tuning using $\lambda = 10$, we also tested $\lambda=0$. The intuition behind using $\lambda=0$ was \textit{``what if we fine-tune w/o a surrogate, ie. $\lambda=0$?"}. In both cases, we finetuned the baseline model for equal number of epochs. 

\textbf{Decision-making task}.  Our real-world experiment focuses on a critical step of the medical decision process:  making the right diagnosis.%\footnote{\BS{While diagnosis is not itself a decision-making task, errors at this stage inevitably degrade subsequent decisions.} \ST{did the reviewers complain about this? If not, I'd avoid focusing on this.}} 
The task is to diagnose $14$ different pathologies (see \cref{tab:results-q2} for the full list) from X-ray images.
In our experiments, we pre-train the \RTWOGEN and \RTWOGENCNM VLMs to predict \textit{findings} from images, and then fine-tune them with \method to improve their generated guidance.  Given that the \textit{impression} is the opinion that the expert forms about potential pathologies visible in the image, \textit{we fine-tune our VLMs to produce textual guidance that -- once interpreted by a human expert -- leads to the same diagnosis entailed by the impression}.

\textbf{Simulating the human expert}.  As in the case of \clevr task, we simulate human decisions using a machine learning model denoted \SurrHuman, for reproducibility.
Specifically, \SurrHuman takes a scan $\vx$ and a corresponding VLM-generated report and diagnoses the $14$ candidate pathologies using three classes:  definitely present (\textit{positive}), definitely absent (\textit{negative}), and unclear (\textit{ambiguous}).
Following \citep{lovelace-mortazavi-2020-learning}, we implement \SurrHuman as a classification model that takes both reports and images as inputs and train it on ground-truth labels obtained by applying the \CHEXPERT \citep{irvin2019chexpert} automated annotation tool to the ground-truth \textit{impressions}. 
% These ground-truth labels act as a proxy to the human decision-maker. However, while using the \CHEXPERT beased labels, we observed that there were many missing labels. Given the fact that there was not mention of a particular label for a particular pathology, we assumed the pathology is absent and therefore we merged all the missing and absents as a single identity -- definitely absent (\textit{negative}).
%
Please note that while the \SurrQuality surrogate is an integral part of \method, \SurrHuman is an experimental detail necessary for evaluation.

In order to emulate a setting with sparse human supervision, we assume \textit{ground-truth labels are available for 10\% of the training data only}. This ground-truth dataset $\calD_{\mathrm{surr}}$ is used for training both the model simulating human decisions \SurrHuman and the quality surrogate model \SurrQuality estimating the quality of these decisions.
We rely on a stratified sampling procedure to select $\calD_{\mathrm{surr}}$ so as to maintain a reasonable coverage of the different classes.

Overall, we proceed as follows.  First, \SurrHuman is trained on $\calD_{\mathrm{surr}}$ to output a diagnosis given ground-truth findings (as these are the only ones for which we know the corresponding human diagnosis).
Once trained, we use \SurrHuman to produce quality ratings by computing the correctness of its predictions over $\calD_{\mathrm{surr}}$, which will later on be used for training the surrogate \SurrQuality.

To avoid biasing the quality rating supervision by computing it on training instances, we run $k$-fold cross validation on $\calD_{\mathrm{surr}}$ and collect quality ratings from the $k$ validation folds. 
For each validation fold, we compute decisions using both VLM-generated guidance and ground-truth text, so as to provide examples of both predicted and ground-truth guidance to train the quality surrogate model.

% \DB{For the systhetic experiment, The \SurrHuman is responsible for evaluating whether the output generated by the VLM satisfies a given rule. It takes as input the hidden representations from the VLM decoder (and optionally the extracted image features) and produces as output a decision indicating rule satisfaction. Internally, the validator generates \textit{logits}, which represent the internal certainty of its decision. These logits are used not only to determine the final boolean outcome but also as input to the subsequent \SurrQuality.}

\textbf{Quality surrogate model}.  As explained in \cref{sec:method-surrogate}, the surrogate \SurrQuality should estimate the quality of human decisions when fed with the VLM guidance. In this experiment, human decisions are proxied with the $14$ labels output by \SurrHuman. The surrogate is thus trained to predict the \textit{correctness} of each of the $14$ predictions made by \SurrHuman. Just like \SurrHuman, the surrogate \SurrQuality used by \method is also implemented as mutli-modal architecture and trained to minimize the average cross entropy loss on the ground-truth dataset described in the previous paragraph. 
% \DB{The synthetic experiment involved \SurrQuality slightly different in nature. The \SurrQuality is trained as a regression model using the Mean Squared Error (MSE) loss function, where the training targets are derived from the \SurrHuman prediction with respect to the ground-truth rule labels. This allows the \SurrQuality to learn patterns that correlate with the \SurrHuman reliability, effectively capturing when the \SurrHuman is likely to be correct or uncertain.}
%
Albeit having different purposes, \SurrHuman and \SurrQuality share the same model architecture. The multimodal functionality of both the models are established with a visual-encoder module and a text encoder module, where the former is a ResNet 101 based model that extracts the features of the radiology images and the latter is a transformer based module that extracts nuanced features of the reports. To this end, we concatenate the features obtained from the text extractor and image extractor before applying a fully connected layer onto the concatenation.

\begin{figure}[!t]
    \centering
    \includegraphics[width=6.6cm]{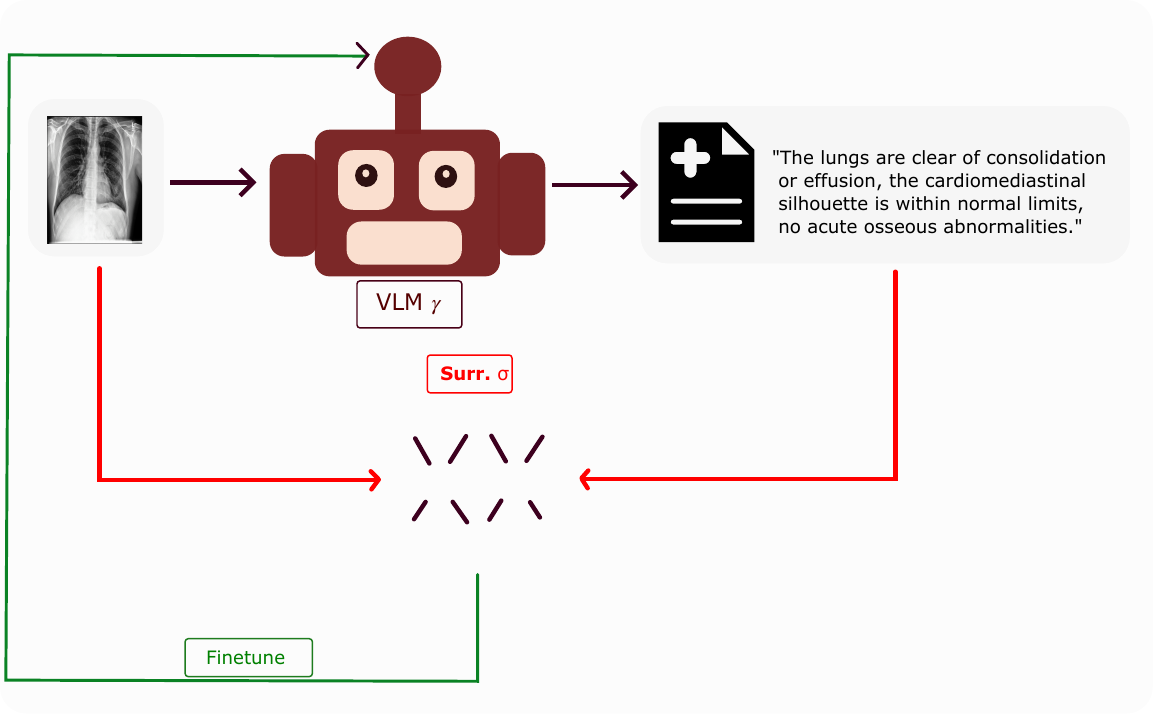}
    \caption{Finetuning policy of \method.}
    \label{fig:ftslog}
\end{figure}

\textbf{Overall pipeline}. First, an \RTWOGEN VLM is pre-trained on the training split $\datatr$ to generate findings.  The VLM is then applied to the decision ground-truth dataset $\calD_{\mathrm{surr}}$ ($10\%$ of the training split). The generated guidance is fed to \SurrHuman to obtain (simulated) human decisions and corresponding quality annotations.  This information is then used to fit the quality surrogate model \SurrQuality.  Finally, the VLM is fine-tuned \textit{on the entire training split}  according to \cref{eq:fine-tune-llm}
%(as the loss for quality surrogate does not require ground-truth supervision) \ST{feels superfluous, it should be clear by now}
for $10$ epochs and evaluated on the test data. The source code is provided in the supplementary material. \cref{fig:ftslog} depicts a summary of the overall pipeline.
%\AP{I added this paragraph for clarity, let's decide whether to keep it} \ST{let's keep it! it helps *a lot*}

% \textbf{Synthetic experiment.}  \ST{@DB: WRITEME}

% \textbf{Medical diagnosis.}  \ST{@DB: WRITEME}

\textbf{Training the VLM}. We trained our baseline VLM with an Nvidia A100 80GB GPU and with batch size 256. We restricted the maximum sequence length to 70 in order to avoid computational over head in later stages of our experiment. We fine-tuned the baseline models (\RTWOGEN and \RTWOGENCNM) on the same GPU with a batch size of 64. The values of all hyper-parameters were taken verbatim from \citep{chen2020generating} and \citep{chen2021cross}, respectively. While training the baseline model, we used a patience of 20 and stored the best model with the highest BLEU-4 score.

\textbf{Training the surrogate}.  In order to emulate a setting with sparse human supervision, we assume \textit{ground-truth labels are available for 10\% of the training data only}. This ground-truth dataset $\calD_{\mathrm{surr}}$ is used for training both the model simulating human decisions \SurrHuman \textit{and} the quality surrogate \SurrQuality estimating the quality of these decisions.
\SurrHuman, which acts as a proxy for the human annotator, is a multimodal classification model which takes the radiology image and corresponding report as inputs and predicts the $14$ target symptoms (cf. \cref{tab:results-q2}).
To this end, in addition to calculating the loss and classification metrics, we conducted a sample-wise comparison between the ground truth labels and the prediction with the purpose of generating training data for \SurrQuality. This comparison yielded an $m\times n$ matrix $Y_q$ where $m = 14$ and $n$ is the number of training examples. Let us consider $G$ is the ground truth matrix of labels used for \SurrHuman and $P$ is the matrix of labels predicted by the model on the validation data. We define, 
\[
\label{quality}
Y_{q} = y_{ij}, \quad \text{where} \ y_{ij}=
\begin{cases}
1&\text{if $G_{ij}=P_{ij}$}\\
0&\text{Otherwise}\\
\end{cases}
\]
In order to train \SurrHuman, we use k-fold cross validation with $k=5$. We assessed the performance of our model on the validation set by scrutinizing the micro $F_1$ score pertaining to the positive mentions.

The \SurrQuality takes same input as the \SurrHuman, but instead of three labels, it outputs either $1$ or $0$ for the $14$ classes (see Equation \ref{quality}). In Table \ref{tab:surrquality_combined}, we report the results obtained from the test split that was used to evaluate the performance of \SurrQuality. Results clearly indicate that \SurrQuality is capable of reliably predicting the correctness of \SurrHuman when provided image and guidance.

\begin{table*}[htbp]
\centering
\caption{\textbf{Outcomes from the test split used to evaluate \SurrQuality.} Results of \RTWOGEN and \RTWOGENCNM  showing per-class, macro, and micro averaged precision ($Pr$), recall ($Rc$), and $F_1$ scores.}
\label{tab:surrquality_combined}
\begin{tabular}{lcccccc}
\toprule
\multirow{2}{*}{\sc Pathology} & \multicolumn{3}{c}{\RTWOGEN} & \multicolumn{3}{c}{\RTWOGENCNM} \\
\cmidrule(lr){2-4} \cmidrule(lr){5-7}
 & $Pr$ & $Rc$ & $F_1$ & $Pr$ & $Rc$ & $F_1$ \\
\midrule
No Findings & 54.88 & 87.53 & 67.46 & 86.22 & 79.69 & 82.82 \\
Cardiomediastinum & 84.68 & 93.47 & 88.85 & 92.65 & 93.24 & 92.94 \\
Cardiomegali & 90.24 & 94.11 & 92.13 & 94.25 & 94.65 & 94.45 \\
Lung Lesion & 78.28 & 93.07 & 85.03 & 91.93 & 91.89 & 91.91 \\
Lung Opacity & 92.07 & 94.21 & 93.13 & 93.61 & 95.48 & 94.54 \\
Edema & 92.52 & 93.87 & 93.19 & 94.99 & 96.82 & 95.90 \\
Consolidation & 91.99 & 92.71 & 92.35 & 93.12 & 94.65 & 93.88 \\
Pneumonia & 42.08 & 84.86 & 56.26 & 74.99 & 75.51 & 75.25 \\
Atelectasis & 82.87 & 94.67 & 88.38 & 93.58 & 92.36 & 92.97 \\
Pneumothorax & 60.89 & 91.55 & 73.13 & 87.55 & 85.46 & 86.49 \\
Pleural Effusion & 96.55 & 97.12 & 96.84 & 96.73 & 98.38 & 97.55 \\
Pleural Other & 62.86 & 87.97 & 73.33 & 84.00 & 81.90 & 82.94 \\
Fracture & 93.94 & 93.86 & 93.90 & 93.62 & 94.36 & 93.99 \\
Support Devices & 78.03 & 91.75 & 84.34 & 91.46 & 88.19 & 89.80 \\
\midrule
\sc MACRO & 92.96 & 80.84 & 86.19 & 90.62 & 90.18 & 90.39 \\
\sc MICRO & 92.20 & 78.71 & 84.17 & 91.18 & 90.84 & 91.01 \\
\bottomrule
\end{tabular}
\end{table*}

\textbf{Applying {\sf \method}}.  We apply \method finetuning for 10 epochs. In this phase, we freeze both the \SurrQuality and the visual encoder layer of the baseline \RTWOGEN model. We try with varying values of $\lambda$ and the best model was chosen based on the validation $F_1$ score. Eventually, for \RTWOGEN, we choose $10$ as the value of hyperparameter $\lambda$ as $\lambda = 10$ yielded the best $F_1$ during finetuning. Along with finetuning using $\lambda = 10$, we also experiment with $\lambda=0$ to finetune without the \SurrQuality. In both cases, we finetune the baseline model for equal number of epochs. We follow the same pipeline for \RTWOGENCNM and chose $0.01$ as the value for hyperparameter $\lambda$. 
\begin{table}[!t]
    \caption{\textbf{{\sc \method} boosts estimated quality of generated guidance without compromising text quality}. The results show that \method substantially improves estimated guidance quality as measured by the surrogate model ($\SurrQuality$) without affecting text quality as measured by BLEU scores over ground-truth caption data.}
    \label{tab:results-q1}
    \centering
    \begin{tabular}{llccccccc}
        \toprule
        {\sc Model} & {\sc Setting}
        & {\sc Bleu$_1$}
        & {\sc Bleu$_2$}
        & {\sc Bleu$_3$}
        & {\sc Bleu$_4$}
        & {\sc BleuRT}
        
        & $\SurrQuality$
        \\
        \midrule
        \multirow{3}{*}{\RTWOGEN} 
            & Pretrained
            & \textbf{0.36} & \textbf{0.22} & \textbf{0.15} & \textbf{0.11} & \textbf{-0.38}  & 0.39 \\
            & Fine-tuned
            & 0.33 & 0.21 & 0.14 & 0.10 & -0.40  & 0.39 \\
            & \method
            & 0.35 & \textbf{0.22} & \textbf{0.15} & \textbf{0.11} & \textbf{-0.38}   & \textbf{0.44} \\
        \midrule
        \multirow{3}{*}{\RTWOGENCNM} 
            & Pretrained
            & \textbf{0.38} & \textbf{0.23} & \textbf{0.16} & \textbf{0.11} & -0.36  & 1.84  \\
            & Fine-tuned
            & 0.37 & 0.22 & 0.15 & \textbf{0.11} & -0.35  & 1.85 \\
            & \method
            & \textbf{0.38} & \textbf{0.23} & \textbf{0.16} & \textbf{0.11} & \textbf{-0.34}  & \textbf{2.02} \\
        \midrule
        \multirow{3}{*}{{\RTWOGENMAMBA}} 
            & {Pretrained}
            & {0.34} & {{0.21}} & {0.15} & \textbf{{0.11}} & {-0.46}  & {2.10}  \\
            & {Fine-tuned}
            & {0.35} & {0.22} & {0.15} & \textbf{{0.11}} & {-0.38}  & {2.11} \\
            & {\method}
            & \textbf{{0.35}} & \textbf{{0.22}} & \textbf{{0.15}} & \textbf{{0.11}} & \textbf{{-0.37}}  & \textbf{{2.12}} \\
        \bottomrule
        \end{tabular}
\end{table}

\begin{figure*}[h]
    \centering
    \includegraphics[width=1.0\linewidth]{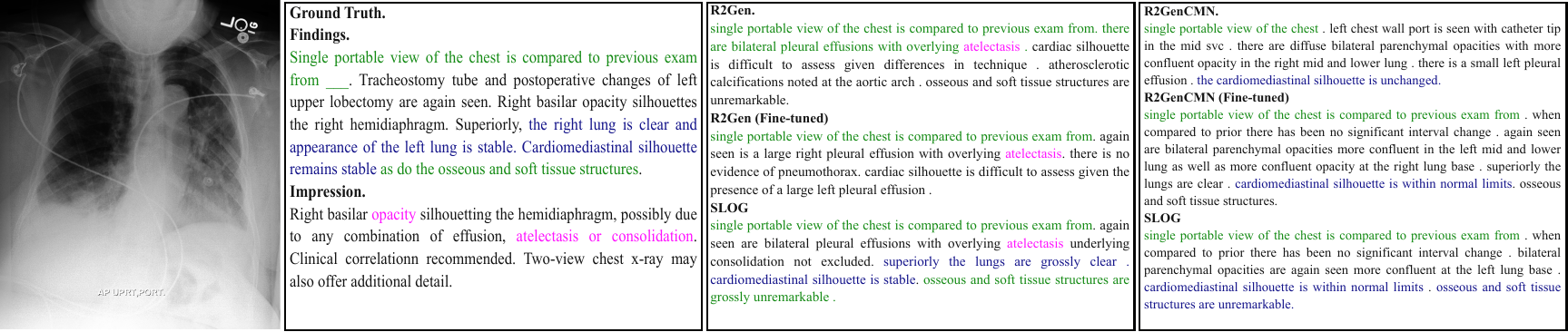}
    \caption{A qualitative example of the improvement of the guidance from \method with respect to the competitors. Green text indicates sentences that are (approximately) shared between the ground truth, \method and at least one of the competitors.  Blue text indicates sentences shared between the ground truth findings (resp. impression) and \method, but missed by the competitors. No ground-truth sentences are shared between ground-truth text and competitors but missed by \method in this example.}
    \label{fig:example-test}
\end{figure*}

\textbf{Q2: {\sf \method} improves informativeness on the test set without compromising BLEU score.} 
%\ST{maybe the simplest way of writing this answer is 1) discuss the results in general, mention they show the surrogate can properly capture the gist of the quality annotations and end-to-end fine-tuning transfers those to the VLM; 2) at the end, explain that the results also mean that the VLM is not overfitting the surrogate.}
%
%\ST{only makes sense if the surrogate does not take $\vx$ as input -- this conflicts with sect 3.  also maybe move to sect 3:} 
A potential issue with using a surrogate model as a proxy of decision quality is that the fine-tuned VLM might end up overfitting the surrogate and produce guidance that, while seemingly informative, is unrelated to the actual input scan.
\method prevents this by complementing estimated guidance quality as computed by the surrogate model with guidance appropriateness for the input image as measured by cross entropy over a training set of findings.
\cref{tab:results-q1} confirms the effectiveness of this strategy.  \method substantially improves estimated guidance quality (second term in \cref{eq:fine-tune-llm}) without compromising text quality, as measured both in terms of BLEU and BLEURT scores over test examples. For the sake of fairness, we compared \method (with $\lambda=10$) with both the pre-trained \RTWOGEN model (before the fine-tuning stage), and the \RTWOGEN model fine-tuned for the same number of epochs as \method, but with caption-level supervision only (i.e., setting $\lambda=0$ in \cref{eq:fine-tune-llm}), as well as with two \RTWOGENCNM and {\RTWOGENMAMBA models, respectively} fine-tuned in the same way. \cref{fig:example-test} shows a qualitative example of the improvement in guidance of \method with respect to the competitors. First, \method's guidance retains all pieces of text that any of the other approach shares with the ground-truth text (green text). On top of this, \method retrieves additional chunks of text that are shared with ground truth findings (blue text) and impression (magenta text), even if the latter are never explicitly included as training supervision, confirming the effectiveness of the quality surrogate in encouraging the generation of relevant guidance for the diagnosis.\footnote{Notice that while the \method guidance text is longer than the one of the competitors in this example, with various chunks of text which are not obviously connected to ground-truth ones, its overall quality is still much higher.}

\begin{table}[!p]
    \centering
    \caption{{Performance of \RTWOGEN finetuned with \method. Results show per-class, macro and micro averaged precision, recall and $F_1$. Best $F_1$ results are boldfaced.}}
    \label{tab:results-q2}
    \scalebox{0.95}{
    \begin{tabular}{lrrrrrrrrr}
        \toprule
         & \multicolumn{3}{c}{{\RTWOGEN}} & \multicolumn{3}{c}{{\RTWOGEN (fine-tuned)}} & \multicolumn{3}{c}{{\method}} \\
         \cmidrule(lr){2-4}   \cmidrule(lr){5-7}   \cmidrule(lr){8-10}
         \sc {Pathology} & {$Pr$} & {$Rc$} & {$F_1$} & {$Pr$} & {$Rc$} & {$F_1$} & {$Pr$} & {$Rc$} & {$F_1$}\\
        
        \midrule
        {No Finding}                 & {37.94} & {65.84} & {48.14} & {38.82} & {65.61} & {\textbf{48.78}} & {39.74} & {62.22} & {48.50} \\
        {Cardiomediastinum}          & {0.0}   & {0.0}   & {0.0}   & {0.0}   & {0.0}   & {0.0}   & {0.0}   & {0.0}   & {0.0}   \\
        {Cardiomegaly}               & {18.49} & {37.77} & {24.83} & {17.21} & {39.36} & {23.95} & {19.49} & {36.70} & {\textbf{25.46}} \\
        {Lung Lesion}                & {0.0}   & {0.0}   & {0.0}   & {0.0}   & {0.0}   & {0.0}   & {0.0}   & {0.0}   & {0.0}   \\
        {Lung Opacity}               & {31.58} & {13.20} & {18.62} & {36.31} & {14.91} & {21.14} & {33.68} & {15.65} & {\textbf{21.37}} \\
        {Edema}                      & {51.45} & {22.83} & {31.63} & {54.47} & {21.54} & {30.88} & {53.44} & {22.51} & {\textbf{31.67}} \\
        {Consolidation}              & {0.0}   & {0.0}   & {0.0}   & {2.44}  & {1.41}  & {1.79}  & {6.25}  & {2.82}  & {\textbf{3.88}}  \\
        {Pneumonia}                  & {0.0}   & {0.0}   & {0.0}   & {0.0}   & {0.0}   & {0.0}   & {50.00} & {0.65}  & {\textbf{1.29}}  \\
        {Atelectasis}                & {21.45} & {27.43} & {24.08} & {21.14} & {27.88} & {24.05} & {21.57} & {32.74} & {\textbf{26.01}} \\
        {Pneumothorax}               & {0.0}   & {0.0}   & {0.0}   & {26.67} & {13.33} & {\textbf{17.78}} & {23.53} & {13.33} & {17.02} \\
        {Pleural Effusion}           & {66.82} & {38.74} & {49.04} & {68.54} & {40.11} & {50.61} & {62.91} & {52.20} & {\textbf{57.06}} \\
        {Pleural Other}              & {0.0}   & {0.0}   & {0.0}   & {0.0}   & {0.0}   & {0.0}   & {0.0}   & {0.0}   & {0.0}   \\
        {Fracture}                   & {0.0}   & {0.0}   & {0.0}   & {0.0}   & {0.0}   & {0.0}   & {0.0}   & {0.0}   & {0.0}   \\
        {Support Devices}            & {28.24} & {54.55} & {37.21} & {28.40} & {53.41} & {37.08} & {28.41} & {55.68} & {\textbf{37.62}} \\
        \midrule
        \sc {MACRO}                  & {18.28} & {18.60} & {16.68} & {21.00} & {19.83} & {18.29} & {24.22} & {21.04} & {\textbf{19.28}} \\
        \sc {MICRO}                  & {33.03} & {31.40} & {32.19} & {33.17} & {31.96} & {32.55} & {33.98} & {33.84} & {\textbf{33.91}} \\
        \bottomrule
    \end{tabular}
    }
\end{table}

\begin{table}[!p]
    \centering
    \caption{{Performance of \RTWOGENCNM finetuned with \method. Results show per-class, macro and micro averaged precision, recall and $F_1$. Best $F_1$ results are boldfaced.}}
    \label{tab:slog-r2gencmn}
    \scalebox{0.95}{
    \begin{tabular}{lrrrrrrrrr}
        \toprule
         & \multicolumn{3}{c}{{\RTWOGENCNM}} & \multicolumn{3}{c}{{\RTWOGENCNM (fine-tuned)}} & \multicolumn{3}{c}{{\method}} \\
         \cmidrule(lr){2-4}   \cmidrule(lr){5-7}   \cmidrule(lr){8-10}
       \sc {Pathology} & {$Pr$} & {$Rc$} & {$F_1$} & {$Pr$} & {$Rc$} & {$F_1$} & {$Pr$} & {$Rc$} & {$F_1$}\\
        
        \midrule
        {No Finding}               & {38.19} & {48.64} & {42.79} & {40.24} & {54.07} & {\textbf{46.14}} & {38.72} & {50.45} & {43.81} \\
        {Cardiomediastinum}          & {0.0}   & {0.0}   & {0.0 }  & {0.0}   & {0.0}   & {0.0}   & {0.0}   & {0.0}   & {0.0}   \\
        {Cardiomegaly}               & {16.33} & {55.85} & {25.27} & {18.30} & {59.57} & {\textbf{28.00}} & {17.73} & {53.19} & {26.60} \\
        {Lung Lesion}                & {0.0}   & {0.0}   & {0.0}   & {0.0}   & {0.0}   & {0.0}   & {33.33} & {1.67}  & {\textbf{3.17}}  \\
        {Lung Opacity}               & {29.55} & {27.38} & {28.43} & {30.74} & {19.32} & {23.72} & {33.81} & {29.10} & {\textbf{31.27}} \\
        {Edema}                      & {48.33} & {9.32}  & {15.63} & {48.21} & {17.36} & {25.53} & {54.03} & {21.54} & {\textbf{30.80}} \\
        {Consolidation}              & {15.71} & {15.49} & {15.60} & {12.50} & {16.90} & {14.37} & {14.61} & {18.31} & {\textbf{16.25}} \\
        {Pneumonia}                  & {0.0}   & {0.0}   & {0.0}   & {15.38} & {2.61}  & {\textbf{4.47}}  & {14.29} & {1.96}  & {3.45}  \\
        {Atelectasis}                & {17.60} & {30.53} & {22.33} & {22.12} & {31.42} & {\textbf{25.96}} & {18.52} & {30.97} & {23.18} \\
        {Pneumothorax}               & {16.67} & {10.00} & {12.50} & {30.00} & {10.00} & {15.00} & {28.57} & {13.33} & {\textbf{18.18}} \\
        {Pleural Effusion}           & {71.43} & {34.34} & {46.38} & {63.90} & {35.99} & {46.05} & {64.44} & {39.84} & {\textbf{49.24}} \\
        {Pleural Other}              & {0.0}   & {0.0}   & {0.0}   & {0.0}   & {0.0}   & {0.0}   & {0.0}   & {0.0}   & {0.0}   \\
        {Fracture}                   & {0.0}   & {0.0}   & {0.0}   & {0.0}   & {0.0}   & {0.0}   & {0.0}   & {0.0}   & {0.0}   \\
        {Support Devices}           & {25.05} & {68.18} & {36.64} & {27.27} & {64.77} & {\textbf{38.38}} & {26.19} & {65.91} & {37.48} \\
        \midrule
        {\sc {MACRO}}                  & {19.92} & {21.41} & {17.54} & {22.05} & {22.29} & {19.12} & {24.59} & {23.31} & {\textbf{20.25}} \\
        {\sc {MICRO}}                  & {27.77} & {31.52} & {29.53} & {30.08} & {32.72} & {31.34} & {29.98} & {34.40} & {\textbf{32.04}} \\
        \bottomrule
    \end{tabular}
    }
\end{table}

\begin{table}[!p]
    \centering
    \caption{{Performance of \RTWOGENMAMBA finetuned with \method. Results show per-class, macro and micro averaged precision, recall and $F_1$. Best $F_1$ results are boldfaced.}}
    \label{tab:rtwogenmamba-results}
    \scalebox{0.95}{
    \begin{tabular}{lrrrrrrrrr}
        \toprule
         & \multicolumn{3}{c}{{\RTWOGENMAMBA}} & \multicolumn{3}{c}{{\RTWOGENMAMBA (fine-tuned)}} & \multicolumn{3}{c}{{\method}} \\
         \cmidrule(lr){2-4}   \cmidrule(lr){5-7}   \cmidrule(lr){8-10}
        \sc {Pathology} & {$Pr$} & {$Rc$} & {$F_1$} & {$Pr$} & {$Rc$} & {$F_1$} & {$Pr$} & {$Rc$} & {$F_1$}\\
        
        \midrule
        {No Finding}                 & {32.97} & {82.81} & {47.16} & {37.36} & {69.23} & {48.53}          & {42.02} & {60.18} & {\textbf{49.49}} \\
        {Enlarged Cardiomediastinum} & {2.63}  & {11.11} & {\textbf{4.26}} & {0.00}  & {0.00}  & {0.00}           & {0.00}  & {0.00}  & {0.00}   \\
        {Cardiomegaly}               & {15.32} & {9.04}  & {11.37} & {19.10} & {27.13} & {22.42}          & {19.38} & {43.09} & {\textbf{26.73}} \\
        {Lung Lesion}                & {0.00}  & {0.00}  & {0.00}  & {0.00}  & {0.00}  & {0.00}           & {0.00}  & {0.00}  & {0.00}   \\
        {Lung Opacity}               & {33.33} & {1.71}  & {3.26}  & {29.49} & {5.62}  & {9.45}           & {24.42} & {10.27} & {\textbf{14.46}} \\
        {Edema}                      & {62.75} & {10.29} & {17.68} & {67.86} & {12.22} & {20.71}          & {50.43} & {18.65} & {\textbf{27.23}} \\
        {Consolidation}              & {8.33}  & {2.82}  & {4.21}  & {9.09}  & {1.41}  & {2.44}           & {5.48}  & {5.63}  & {\textbf{5.56}}  \\
        {Pneumonia}                  & {0.00}  & {0.00}  & {0.00}  & {0.00}  & {0.00}  & {0.00}           & {0.00}  & {0.00}  & {0.00}   \\
        {Atelectasis}                & {25.62} & {13.72} & {17.87} & {23.83} & {22.57} & {23.18}          & {22.13} & {36.73} & {\textbf{27.62}} \\
        {Pneumothorax}               & {0.00}  & {0.00}  & {0.00}  & {15.38} & {13.33} & {14.29}          & {21.05} & {13.33} & {\textbf{16.33}} \\
        {Pleural Effusion}           & {66.84} & {34.34} & {45.37} & {58.30} & {45.33} & {51.00}          & {56.62} & {55.22} & {\textbf{55.91}} \\
        {Pleural Other}              & {0.00}  & {0.00}  & {0.00}  & {0.00}  & {0.00}  & {0.00}           & {0.00}  & {0.00}  & {0.00}   \\
        {Fracture}                   & {0.00}  & {0.00}  & {0.00}  & {0.00}  & {0.00}  & {0.00}           & {0.00}  & {0.00}  & {0.00}   \\
        {Support Devices}            & {29.35} & {48.86} & {36.67} & {25.77} & {52.27} & {34.52}          & {26.47} & {66.48} & {\textbf{37.86}} \\
        \midrule
        {\sc MACRO}                  & {19.80} & {15.34} & {13.42} & {20.44} & {17.79} & {16.18}          & {19.14} & {22.11} & {\textbf{18.66}} \\
        {\sc MICRO}                  & {33.50} & {26.69} & {29.71} & {32.96} & {29.20} & {30.97}          & {31.69} & {34.20} & {\textbf{32.90}} \\
        \bottomrule
    \end{tabular}
    }
\end{table}

\textbf{Q3: {\sf \method} improves quality of decisions.} \cref{tab:results-q2,tab:slog-r2gencmn,tab:rtwogenmamba-results} show the results in terms of decision quality, as measured by the $F_1$ score of the positive label for all 14 classes (multi-label prediction).
Results clearly indicate the effectiveness of the \method guidance in improving decision quality, despite the modest amount of supervision it received.
 On \RTWOGEN, \method outperforms the competitors in 8 out of 14 classes. All methods fail to identify any positive occurrence for three particularly under-represented classes (\texttt{Pneumonia}, \texttt{Pleural Other}, \texttt{Fracture}), while \method slightly under-performs with respect to pre-trained \RTWOGEN in 2 classes only. It is worth noticing that \method is especially effective in improving recall without affecting precision on average, as shown by the two bottom lines reporting results averaged over classes (macro) and instances (micro) respectively. On \RTWOGENCNM, \method is the best performing method on {6 classes, while the fine-tuned baseline wins on 5 classes, leaving the rest 3 as a tie among the three competitors. Additionally \method helps in recovering few remote cases that are missed with both the pretrained and the finetuned models, for example, `Pneumonia' for \RTWOGEN and \RTWOGENCNM. The \method model developed on \RTWOGENMAMBA, yields highest $F_1$ in 9 classes, while for the rest of the 5 classes, all the competitors fail to yield correct predictions. \method eventually yields highest $F1$ both in terms of `micro' and `macro' measures compared to its competitors. We also observe similar performance improve of \method comapared to its competitors for all the three models when evaluated with ChexBert \citep{smit2020combining}. The results are added to the \Cref{tab:results-mamba-comprehensive,tab:results-rgen-chexbert,tab:results-rcmn-chexbert} in \Cref{appendix}.}

\textbf{Q4: {\sf \method} has the potential to help doctors make better decisions.}
% The guidance quality of \method\ was evaluated against a fine-tuned \RTWOGENCNM\ via expert review by a pulmonologist with three years of clinical experience. In this setup, the clinician assessed 25 guidances from \method and fine-tuned \RTWOGENCNM (total 50).
%
% To ensure representative coverage across diagnostic categories while maintaining balanced class counts, we employed stratified sampling with a minimum positive-count constraint. Let \( D \) denote the full dataset of reports and \( \mathcal{C} \) the set of classes (e.g., CheXpert pathology labels). We select \( S \subseteq D \) of size \( N \) such that each class attains at least the required number of positive examples. The sampling algorithm is detailed in \Cref{appendix}.
%In this set-up, a subset of $25$ guidances from each model were provided to the doctor. The doctor has 3 years of experience in pulmonology, a domain in medical science highly suitable for the type of problem that we are addressing.  To ensure representative sampling across multiple diagnostic categories while maintaining balanced class coverage, we employ a stratified sampling strategy with minimum class coverage. Let \( D \) denote the complete dataset of reports, and let \( \mathcal{C} \) represent the set of possible classes (e.g., the CheXpert pathology labels). The goal is to obtain a subset \( S \subseteq D \) of size \( N \), such that each class has at least a minimum number of positive examples. The actual algorithm has been provided in the \Cref{appendix}.
The guidance quality of \method\ was evaluated against a fine-tuned \RTWOGENCNM\ via expert review by a pulmonologist with three years of clinical experience. In this setup, the clinician assessed {55} guidances from \method and fine-tuned \RTWOGENCNM (total 50). We decided to focus on fine-tuned \RTWOGENCNM, the runner-up according to the previous experiment, to avoid overloading the clinician with too many assessments.
It is important to examine whether \method helps the physician identify the presence of a symptom. Therefore, to ensure representative coverage while maintaining balanced class counts, we implemented a sampling technique that guaranties the presence of a symptom with at least one positive mention in the sample set. Thus, the samples cover all classes with at least one positive mention. The actual algorithm has been provided in the \Cref{appendix}. {To this end, the entire task of annotation was randomized and blinded for the physician. In the sense, the physician, received an unordered representation of the reports and received no clue about the model that is responsible for a particular guidance. In order to ensure a fair competition between \method and its competitor, the physician was asked to annotate the same studies for both the models. \cref{tab:slog-doctor} shows the results in terms of decision quality, measured by the $F_1$ score of the positive label for the 14 classes (prediction with multiple labels). The results confirm the advantage of \method in improving the overall performance of the decision-making process, both in terms of micro and macro $F_1$. In order to validate the results from 55 samples, the mean $F_1$ scores for each of the labels has also been reported. While \method falls short on 4 classes including `No finding', `Atelectasis', `Pneumothorax' and `Pleural Effusion', it outperforms the baseline on 7 classes. Additionally, in \Cref{tab:raw-counts-comparison} we examined the pattern of error of the two models based on their confusion matrix transitions (Baseline $\rightarrow$ \method). We focused on the pathologies responsible for largest inflation in error, i.e. $\Delta FP + \Delta FN$.  In particular `Lung lesion' ($+10$ FP), `Atelectasis' ($+6$ TP, $-1$ TP) and Pleural Effusion ($+4$ FP, $-1$ TP). It is worth noticing that the dominant source of error is due to over predicting, rather than missing true pathologies.  Furthermore, rare instances of positive $\Delta$ in \textit{false negatives} bolsters the fact that the lower performance of \method majorly stems from over detection, rather than missing actual cases – a phenomenon often considered to be less risky in the medical domain.}

\begin{table}[!p]
    \centering
    \caption{{Performance comparison between baseline (fine-tuned) and \method when a subset of samples was presented to a doctor. Results show per-class, macro and micro averaged precision, recall and $F_1$. Best $F_1$ results between baseline and \method\ are boldfaced.}}
    \label{tab:slog-doctor}
    \scriptsize
    \begin{tabular}{lrrrr|rrrr}
        \toprule
         & \multicolumn{4}{c}{{\RTWOGENCNM (Fine-tune)}} & \multicolumn{4}{c}{{\method}} \\
         \cmidrule(lr){2-5}   \cmidrule(lr){6-9}
        \sc {Pathology} & {$Pr$} & {$Rc$} & {$F_1$} & {Bootstrap $F_1$ [95\% CI]} & {$Pr$} & {$Rc$} & {$F_1$} & {Bootstrap $F_1$ [95\% CI]}\\
        
        \midrule
        {No Finding}                 & {23.53} & {44.44} & {\textbf{30.77}} & {29.60 [8.30, 53.30]}  & {8.33}  & {11.11} & {9.52}           & {9.70 [0.00, 28.60]} \\
        {Enlarged Cardiomediastinum} & {0.00}  & {0.00}  & {0.00}           & {0.00 [0.00, 0.00]}    & {0.00}  & {0.00}  & {0.00}           & {0.00 [0.00, 0.00]} \\
        {Cardiomegaly}               & {23.81} & {50.00} & {32.26}          & {31.40 [8.70, 51.90]}  & {32.00} & {80.00} & {\textbf{45.71}} & {45.20 [22.80, 63.20]} \\
        {Lung Lesion}                & {0.00}  & {0.00}  & {0.00}           & {0.00 [0.00, 0.00]}    & {7.14}  & {33.33} & {\textbf{11.76}} & {11.40 [0.00, 33.40]} \\
        {Lung Opacity}               & {66.67} & {17.39} & {27.59}          & {26.60 [6.70, 47.60]}  & {50.00} & {30.43} & {\textbf{37.84}} & {36.90 [17.10, 57.10]} \\
        {Edema}                      & {20.00} & {28.57} & {23.53}          & {21.60 [0.00, 47.60]}  & {25.00} & {57.14} & {\textbf{34.78}} & {34.20 [9.50, 57.10]} \\
        {Consolidation}              & {33.33} & {33.33} & {33.33}          & {27.70 [0.00, 80.00]}  & {37.50} & {100.0} & {\textbf{54.55}} & {52.70 [0.00, 85.70]} \\
        {Pneumonia}                  & {0.00}  & {0.00}  & {0.00}           & {0.00 [0.00, 0.00]}    & {20.00} & {11.11} & {\textbf{14.29}} & {13.10 [0.00, 40.00]} \\
        {Atelectasis}                & {27.27} & {60.00} & {\textbf{37.50}} & {35.80 [14.30, 56.20]} & {18.52} & {50.00} & {27.03}           & {26.80 [6.50, 46.20]} \\
        {Pneumothorax}               & {50.00} & {50.00} & {\textbf{50.00}} & {40.50 [0.00, 100.00]} & {16.67} & {50.00} & {25.00}           & {20.90 [0.00, 66.70]} \\
        {Pleural Effusion}           & {66.67} & {62.50} & {\textbf{64.52}} & {64.60 [40.00, 82.40]} & {50.00} & {56.25} & {52.94}           & {52.20 [30.80, 71.80]} \\
        {Pleural Other}              & {0.00}  & {0.00}  & {0.00}           & {0.00 [0.00, 0.00]}    & {0.00}  & {0.00}  & {0.00}           & {0.00 [0.00, 0.00]} \\
        {Fracture}                   & {0.00}  & {0.00}  & {0.00}           & {0.00 [0.00, 0.00]}    & {0.00}  & {0.00}  & {0.00}           & {0.00 [0.00, 0.00]} \\
        {Support Devices}            & {15.00} & {42.86} & {22.22}          & {21.60 [0.00, 42.90]}  & {27.27} & {85.71} & {\textbf{41.38}} & {40.40 [18.20, 62.50]} \\
        \midrule
        {\sc MACRO}                  & {23.31} & {27.79} & {22.98}          & {---}                  & {20.89} & {40.36} & {\textbf{25.34}} & {---} \\
        {\sc MICRO}                  & {25.00} & {36.00} & {29.51}          & {---}                  & {24.34} & {46.00} & {\textbf{31.83}} & {---} \\
        \bottomrule
    \end{tabular}
    
\end{table}

\begin{table}[!htbp]
\centering
\caption{{Comparison of Raw Diagnostic Counts of the physician's annotations: Baseline vs. \method}}
\label{tab:raw-counts-comparison}
\scalebox{0.8}{
\begin{tabular}{lcccc|cccc}
\toprule
 & \multicolumn{4}{c}{{Baseline}} & \multicolumn{4}{c}{{\method}} \\
 \cmidrule(lr){2-5} \cmidrule(lr){6-9}
{\textbf{Pathology}} & {\textbf{TP}} & {\textbf{FP}} & {\textbf{TN}} & {\textbf{FN}} & {\textbf{TP}} & {\textbf{FP}} & {\textbf{TN}} & {\textbf{FN}} \\
\midrule
{No Finding}                 & {4}  & {13} & {33} & {5}  & {1}  & {11} & {35} & {8} \\
{Enlarged Cardiomediastinum} & {0}  & {17} & {38} & {0}  & {0}  & {14} & {41} & {0} \\
{Cardiomegaly}               & {5}  & {16} & {29} & {5}  & {8}  & {17} & {28} & {2} \\
{Lung Lesion}                & {0}  & {3}  & {49} & {3}  & {1}  & {13} & {39} & {2} \\
{Lung Opacity}               & {4}  & {2}  & {30} & {19} & {7}  & {7}  & {25} & {16} \\
{Edema}                      & {2}  & {8}  & {40} & {5}  & {4}  & {12} & {36} & {3} \\
{Consolidation}              & {1}  & {2}  & {50} & {2}  & {3}  & {5}  & {47} & {0} \\
{Pneumonia}                  & {0}  & {3}  & {43} & {9}  & {1}  & {4}  & {42} & {8} \\
{Atelectasis}                & {6}  & {16} & {29} & {4}  & {5}  & {22} & {23} & {5} \\
{Pneumothorax}               & {1}  & {1}  & {52} & {1}  & {1}  & {5}  & {48} & {1} \\
{Pleural Effusion}           & {10} & {5}  & {34} & {6}  & {9}  & {9}  & {30} & {7} \\
{Pleural Other}              & {0}  & {3}  & {52} & {0}  & {0}  & {5}  & {50} & {0} \\
{Fracture}                   & {0}  & {2}  & {52} & {1}  & {0}  & {3}  & {51} & {1} \\
{Support Devices}            & {3}  & {17} & {31} & {4}  & {6}  & {16} & {32} & {1} \\
\bottomrule
\end{tabular}
}
\end{table}

\begin{figure}[!b]%
    \centering
    \subfloat[\centering Micro $F_1$ on the test data]{{\includegraphics[width=5cm]{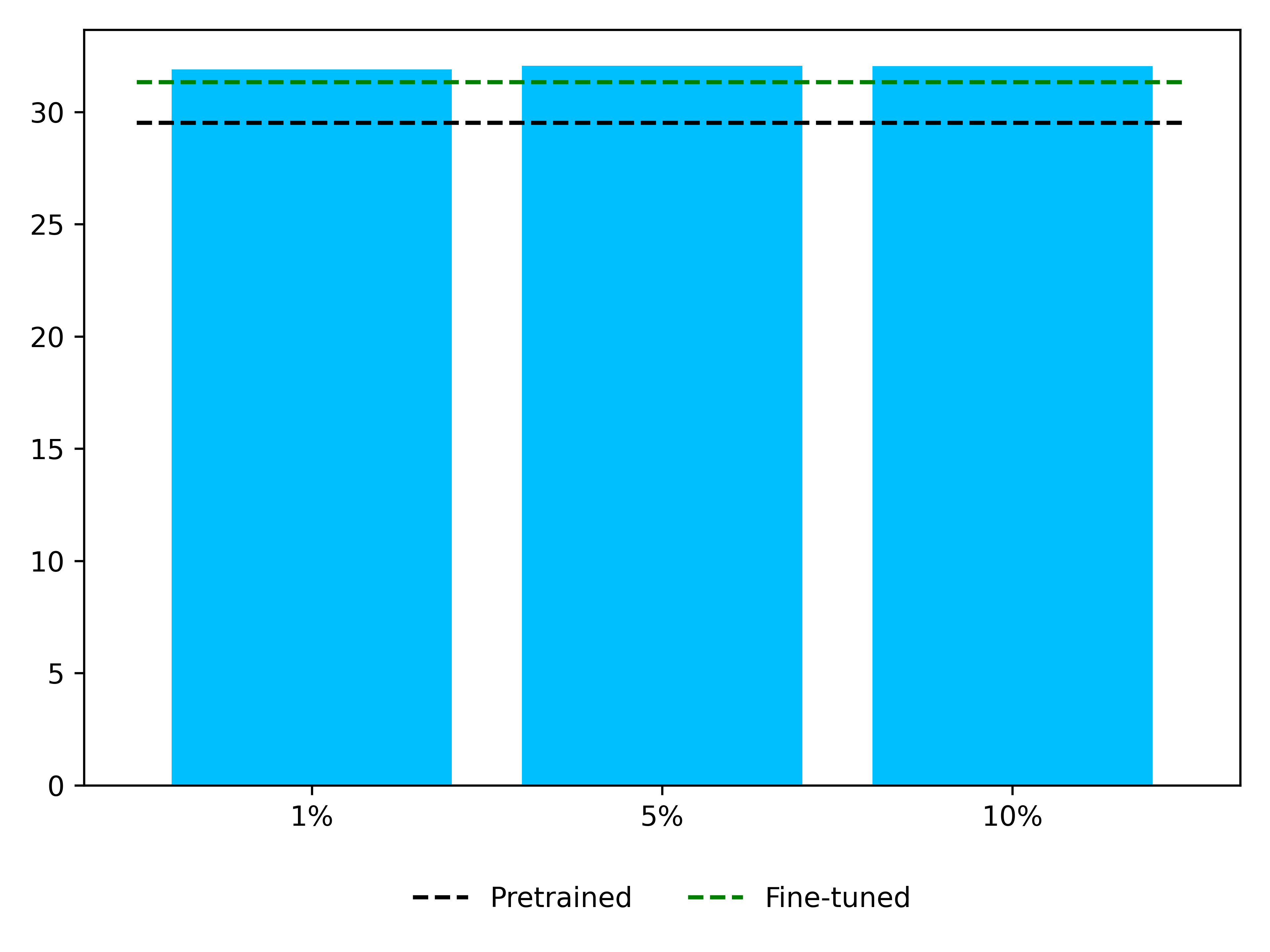} }}%
    \qquad
    \subfloat[\centering Macro $F_1$ on the test data]{{\includegraphics[width=5cm]{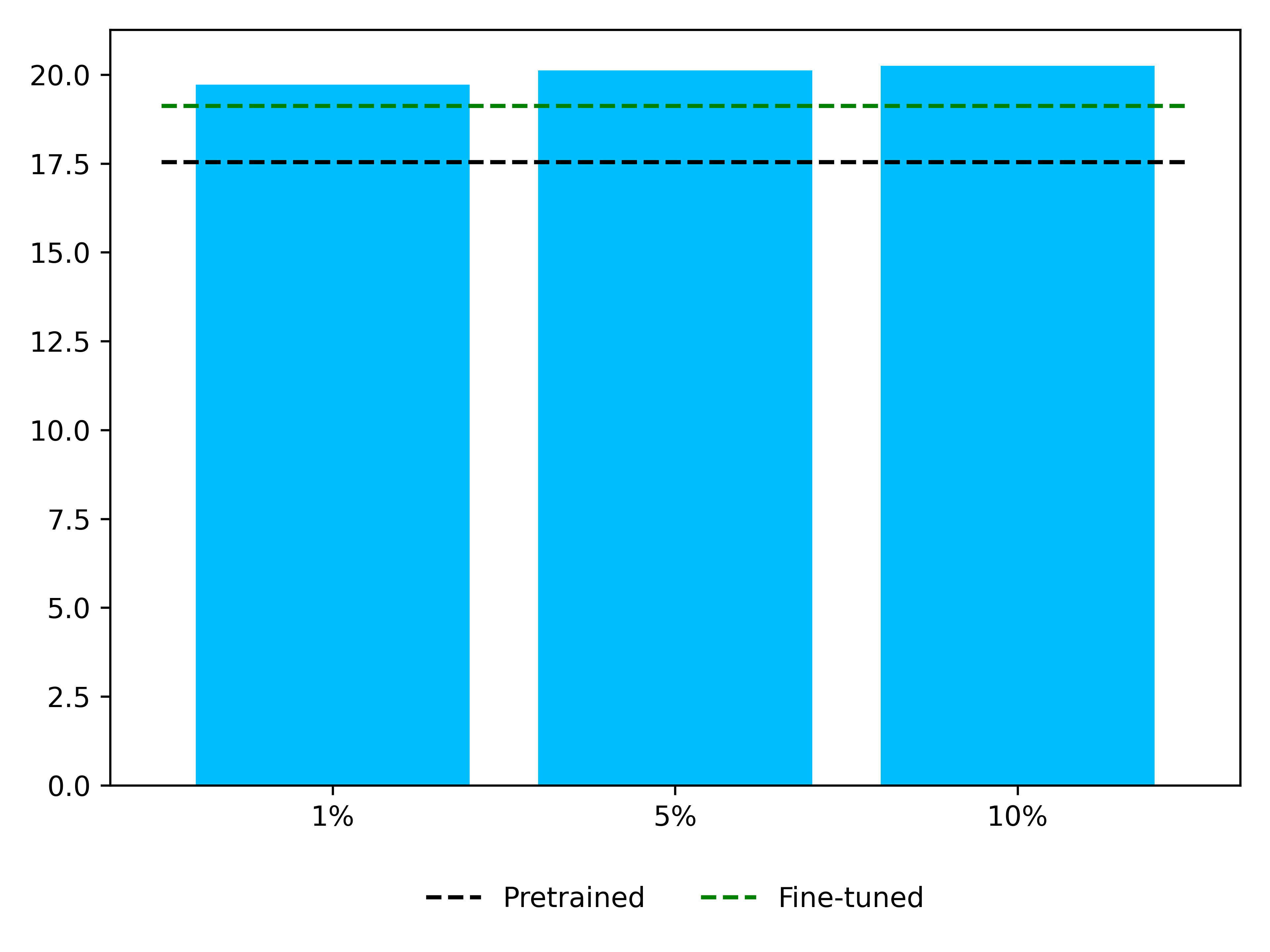} }}%
    \caption{Performance of the finetuned models involving \SurrQuality trained with varying percentage of train examples. The black and green lines are the $F_1$ scores obtained with the pretrained and finetuned models ($\lambda = 0$).}%
    \label{fig:ablation-R2Gencmn}%
\end{figure}

\textbf{Ablation experiment.}  To investigate the influence of \SurrQuality in the finetuning process, we conduct an ablation study based on the percentage of training data used to train \SurrHuman and \SurrQuality. In our primary experiments, we take $10\%$ of training examples to train the surrogate models. However, for the ablation, we finetune the baseline model \SurrQuality with only $1\%$ and $5\%$ of the training examples.
The results can be viewed in \cref{fig:ablation-R2Gencmn}.  While the surrogates trained with $5\%$ of the data performed at par with surrogates trained with $10\%$, an improvement can still be observed when reducing the amount of training data further.  In particular, macro $F_1$ (right) shows that even with $1\%$ surrogate training data, \method still manages to outperform both baselines (dashed lines), while micro $F_1$ (left) is on-par with them.
This indicates that even the less informed surrogates can help improving prediction quality for the rarer -- but possibly significant for decision making -- classes.
In \cref{tab:slog_ablation}, we notice that \method trained with $10\%$ of training examples outperforms the former two in both micro and macro metrics. Detailed results can be obtained in \cref{tab:slog_ablation_full}.

\begin{table}[t]
\centering
\caption{{Ablation study on dataset size for \method.}}
\label{tab:slog_ablation}
\begin{tabular}{llccc}
\toprule
\textbf{{Type}} & \textbf{{Metric}} & \textbf{{\method 1\%}} & \textbf{{\method 5\%}} & \textbf{{\method 10\%}} \\
\midrule
\multirow{3}{*}{{\textbf{Macro}}} 
 & {Precision} & {21.56} & {21.51} & {24.59} \\
 & {Recall}    & {23.15} & {23.54} & {23.31} \\
 & {F$_1$}     & {19.72} & {20.12} & {20.25} \\
\midrule
\multirow{3}{*}{{\textbf{Micro}}} 
 & {Precision} & {30.06} & {30.02} & {29.98} \\
 & {Recall}    & {33.96} & {34.40} & {34.40} \\
 & {F$_1$}     & {31.89} & {32.06} & {32.04} \\
\bottomrule
\end{tabular}
\end{table}

\section{Related Work}
\label{sec:related-work}

\textbf{Aligning LLMs.}  The standard approach for aligning large language models to human interests is reinforcement learning with human feedback (RLHF) \citep{ziegler2020finetuning, ouyang2022training}.
Several RLHF-based approaches that target medical tasks exist. \citet{yunxiang2023chatdoctor} and  \citet{wang2023chatcad} proposed medical chat models obtained by fine-tuning existing architectures, while \citet{Bazi2023} introduced a specially designed vision transformer. \citet{seo2020reinforcing} presented a method for improving the performance of an image caption generator with offline human feedback.
\method can be viewed as a variant of RLHF that foregoes reinforcement learning in favor of a fully end-to-end fine-tuning strategy, for efficiency.  It also differs in aim, in that it optimizes the model's guidance for \textit{a specific human decision making task}, rather than for factuality and fairness in general \citep{ouyang2022training}.

\textbf{Pathological report generation.}  Several approaches \citep{hou2021ratchet, chen2020generating, wang2022cross, wang2023r2gengpt} have been developed for machine-driven pathological report generation from chest X-ray images using the \MimicCXR \citep{johnson2019mimic} and the Indiana University chest X-ray data sets \citep{demner2016preparing}. \citep{lovelace2020learning} designed a model with a similar objective but they proposed to leverage the \CHEXPERT dataset to enhance the coherence of their model. \citep{tanida2023interactive} introduced a region-guided model to generate pathological reports, thus opening the window of interactive human-guided report generation.  \citep{srivastav2024maira} used a large language model based on Vicuna-7B to generate radiology reports out of CXR images. A slightly different approach was used by \citep{woznicki2024automatic} where the authors used a large language model to extract the structured information out of the \textit{findings} section of a report. However, these models are not concerned with optimizing the utility of the generated reports for the follow-up decision making. Our approach builds on top of these methods, enriching them with the ability to incorporate surrogate quality information (we use \citep{chen2020generating} in our evaluation, but any of these models can be adapted for \method).

\textbf{Other approaches.}  LTG is related to \textit{explain then predict} (ETP) \citep{camburu2018snli, kumar2020nile, zhao2021lirex}, a framework for building explainable \citep{guidotti2018survey} models in which a machine first outputs a full-fledged explanation -- playing the role of ``guidance'' -- and then derives a prediction from the explanation itself.  In LTG, however, the prediction step is carried out by a human expert, and as such it is not differentiable.  Also, ETP requires direct supervision on the explanations themselves, which is seldom available.  In contrast, \method improves guidance quality using indirect scoring feedback, which is comparatively easier to acquire.

Finally, LTG is not restricted to textual guidance.  One option is, for instance, to implement guidance in terms of explanations extracted from (or output by) an underlying image classifier to guide human decision making \citep{guidotti2018survey}.
From this perspective, LTG is tied to explanatory interactive learning (XIL) \citep{schramowski2020making, teso2023leveraging}, in which the goal is to improve the quality of explanations output by a machine learning model by interactively acquiring corrections to the explanations themselves.
The key difference is that LTG focuses on down-stream decision quality and \method supports textual guidance, while XIL aims at more generally improving explanation quality and implementations do no support textual explanations.

{\section{Discussion}}
{One of the fundamental properties of \method is it works on the top of a pretrained model. Therefore, the performance of \method varies over the performance of the pretrained model. We also tried to experiment the performance of \method with respect to LLava model. However, at its primary inferential stage, the pretrained LLava model yielded sub-optimal results (13.66 \% Micro $F_1$), possibly because the model is not particularly designed to perform complex medical tasks. Considering the models that have been tried in this study, the comparatively low performance of the llava model restricted us from assessing the performance of \method on the top of Llava as \method cannot dramatically improve the performance of the pretrained model. 
The core purpose of \textit{learning to guide} is to provide guidance for the human decision maker. However, the performance of human may not be uniform and may depend on several other parameters and we reserved the analysis uniformity of impact of AI system on humans for future studies.}

\section{Conclusion}
\label{sec:conclusion}

We introduced \textit{learning to guide} as an alternative setup for high-stakes hybrid decision making that ensures the human expert is always in the loop, as well as \method, an end-to-end approach for turning pre-trained VLMs into high-quality textual guidance using human feedback.
Our results suggest that \method is effective at steering VLMs towards generating more informative guidance, leading to improved accuracy in downstream decisions. 

In follow-up work, we plan to extend \method by integrating ideas from active learning \citep{settles2012active} to acquire the quality rankings, and to explore connections with explainable AI \citep{guidotti2018survey}, explanatory interactive learning \citep{schramowski2020making, teso2023leveraging} and skeptical learning \citep{zeni2019fixing} to facilitate the identification and correction of potential issues with the generated guidance.

\bibliographystyle{plain}
\bibliography{paper,explanatory-supervision}

\appendix
\section{Appendix}
\label{appendix}

\begin{table}[h]
    \centering
    \caption{{Ablation study of \method based on \RTWOGENCNM with different proportions of labeled data (SLOG $1\%$, $5\%$, and $10\%$).}}
    \label{tab:slog_ablation_full}
    \scriptsize
    \begin{tabular}{lrrrrrrrrr}
        \toprule
         & \multicolumn{3}{c}{{\method 1\%}} & \multicolumn{3}{c}{{\method 5\%}} & \multicolumn{3}{c}{{\method 10\%}} \\
         \cmidrule(lr){2-4}   \cmidrule(lr){5-7}   \cmidrule(lr){8-10}
        \sc {Pathology} & {$Pr$} & {$Rc$} & {$F_1$} & {$Pr$} & {$Rc$} & {$F_1$} & {$Pr$} & {$Rc$} & {$F_1$}\\
        
        \midrule
        {No Finding}                 & {38.56} & {49.55} & {43.37} & {39.60} & {49.55} & {44.02} & {38.72} & {50.45} & {43.81} \\
        {Enlarged Cardiomediastinum} & {1.27}  & {5.56}  & {2.06}  & {1.39}  & {5.56}  & {2.22}  & {0.00}  & {0.00}  & {0.00}  \\
        {Cardiomegaly}               & {17.41} & {54.26} & {26.36} & {17.01} & {52.66} & {25.71} & {17.73} & {53.19} & {26.60} \\
        {Lung Lesion}                & {0.00}  & {0.00}  & {0.00}  & {0.00}  & {0.00}  & {0.00}  & {33.33} & {1.67}  & {3.17}  \\
        {Lung Opacity}               & {33.33} & {26.16} & {29.32} & {33.33} & {27.38} & {30.07} & {33.81} & {29.10} & {31.27} \\
        {Edema}                      & {53.44} & {22.51} & {31.67} & {53.28} & {23.47} & {32.59} & {54.03} & {21.54} & {30.80} \\
        {Consolidation}              & {15.66} & {18.31} & {16.88} & {13.98} & {18.31} & {15.85} & {14.61} & {18.31} & {16.25} \\
        {Pneumonia}                  & {14.29} & {1.31}  & {2.40}  & {8.33}  & {1.31}  & {2.26}  & {14.29} & {1.96}  & {3.45}  \\
        {Atelectasis}                & {20.69} & {31.86} & {25.09} & {20.28} & {32.30} & {24.91} & {18.52} & {30.97} & {23.18} \\
        {Pneumothorax}               & {17.65} & {10.00} & {12.77} & {25.00} & {13.33} & {17.39} & {28.57} & {13.33} & {18.18} \\
        {Pleural Effusion}           & {64.22} & {40.93} & {50.00} & {64.02} & {42.03} & {50.75} & {64.44} & {39.84} & {49.24} \\
        {Pleural Other}              & {0.00}  & {0.00}  & {0.00}  & {0.00}  & {0.00}  & {0.00}  & {0.00}  & {0.00}  & {0.00}  \\
        {Fracture}                   & {0.00}  & {0.00}  & {0.00}  & {0.00}  & {0.00}  & {0.00}  & {0.00}  & {0.00}  & {0.00}  \\
        {Support Devices}            & {25.28} & {63.64} & {36.19} & {24.94} & {63.64} & {35.84} & {26.19} & {65.91} & {37.48} \\
        \midrule
        {\sc MACRO}                  & {21.56} & {23.15} & {19.72} & {21.51} & {23.54} & {20.12} & {24.59} & {23.31} & {20.25} \\
        {\sc MICRO}                  & {30.06} & {33.96} & {31.89} & {30.02} & {34.40} & {32.06} & {29.98} & {34.40} & {32.04} \\
        \bottomrule
    \end{tabular}
    
\end{table}

\begin{algorithm}[H]
\caption{Sampling with Minimum Class Coverage}
\KwIn{
  Dataset $D$ (rows = reports), \\
  label set $\mathcal{C}$ (e.g., CheXpert classes), \\
  target sample size $N$ (e.g., $30$), \\
  minimum positives per class $m$ (e.g., $2$)
}
\KwOut{Sampled subset $S \subseteq D$ of size $N$}
$S \gets \emptyset$ \tcp*{selected indices/reports}

\ForEach{$c \in \mathcal{C}$}{
  $P_c \gets \{ i \in D \mid \text{label}(i,c)=1 \}$\;
  \eIf{$|P_c| \ge m$}{
    choose $m$ elements uniformly from $P_c$ without replacement and add to $S$\;
  }{
    add all of $P_c$ to $S$ 
  }
}

\If{$|S| > N$}{
  uniformly sub-sample $S$ down to size $N$; \Return{$S$}\;
}

$R \gets D \setminus S$ \tcp*{remaining pool}
\While{$|S| < N$ \textbf{and} $R \neq \emptyset$}{
  pick $x \in R$ uniformly at random; add $x$ to $S$; remove $x$ from $R$\;
}

\Return{$S$}
\end{algorithm}

\begin{table}[!p]
    \caption{{\textbf{{\sf \method} vs R2Gen-Mamba (Pretrained \& Fine-tuned) evaluated with \textbf{ChexBert}}}}
    \label{tab:results-mamba-comprehensive}
    \centering
    \scalebox{0.85}{
        \begin{tabular}{lrrrrrrrrr}
            \toprule
                & \multicolumn{3}{c}{{R2Gen-Mamba (Pre)}} & \multicolumn{3}{c}{{R2Gen-Mamba (FT)}} & \multicolumn{3}{c}{{\method}} \\
          \cmidrule(lr){2-4} \cmidrule(lr){5-7} \cmidrule(lr){8-10}
         \sc {Pathology} & {$Pr$} & {$Rc$} & {$F_1$} & {$Pr$} & {$Rc$} & {$F_1$} & {$Pr$} & {$Rc$} & {$F_1$} \\
        \midrule
        {Enlarged Card.}    & {7.04}   & {3.36}  & {4.55}  & {6.25}   & {2.68}  & {3.76}  & {15.00} & {6.04}  & {\textbf{8.61}}  \\
        {Cardiomegaly}      & {67.08}  & {26.52} & {38.02} & {60.91}  & {39.54} & {47.95} & {64.45} & {41.52} & {\textbf{50.50}} \\
        {Lung Opacity}      & {42.25}  & {9.65}  & {15.71} & {42.97}  & {8.84}  & {14.67} & {49.58} & {9.49}  & {\textbf{15.92}} \\
        {Lung Lesion}       & {100.00} & {1.06}  & {\textbf{2.11}}  & {100.00} & {1.06}  & {\textbf{2.11}}  & {0.00}  & {0.00}  & {0.00}  \\
        {Edema}             & {46.27}  & {10.92} & {17.66} & {47.93}  & {20.42} & {\textbf{28.64}} & {48.70} & {19.72} & {28.07} \\
        {Consolidation}     & {33.33}  & {2.33}  & {4.35}  & {20.00}  & {2.33}  & {4.17}  & {27.27} & {3.49}  & {\textbf{6.19}}  \\
        {Pneumonia}         & {25.00}  & {4.00}  & {6.90}  & {50.00}  & {8.00}  & {\textbf{13.79}} & {50.00} & {8.00}  & {\textbf{13.79}} \\
        {Atelectasis}       & {31.27}  & {30.66} & {30.96} & {33.07}  & {34.81} & {\textbf{33.92}} & {32.90} & {34.81} & {33.83} \\
        {Pneumothorax}      & {30.77}  & {27.59} & {\textbf{29.09}} & {25.00}  & {17.24} & {20.41} & {26.32} & {17.24} & {20.83} \\
        {Pleural Effusion}  & {75.93}  & {40.94} & {53.20} & {67.61}  & {53.24} & {\textbf{59.57}} & {66.02} & {53.02} & {58.81} \\
        {Pleural Other}     & {0.00}   & {0.00}  & {0.00}  & {0.00}   & {0.00}  & {0.00}  & {0.00}  & {0.00}  & {0.00}  \\
        {Fracture}          & {0.00}   & {0.00}  & {0.00}  & {0.00}   & {0.00}  & {0.00}  & {0.00}  & {0.00}  & {0.00}  \\
        {Support Devices}   & {74.75}  & {41.76} & {53.58} & {75.63}  & {54.58} & {63.40} & {75.99} & {56.23} & {\textbf{64.63}} \\
        {No Finding}        & {9.34}   & {\textbf{85.26}} & {16.84} & {10.14}  & {77.89} & {17.94} & {10.45} & {78.95} & {\textbf{18.45}} \\
        \midrule
        {MACRO}             & {38.79}  & {20.29} & {19.50} & {38.54}  & {22.90} & {22.17} & {33.33} & {\textbf{23.46}} & {\textbf{22.83}} \\
        {MICRO}             & {37.46}  & {24.81} & {29.85} & {42.46}  & {31.42} & {36.12} & {\textbf{43.81}} & {\textbf{32.22}} & {\textbf{37.13}} \\
        \bottomrule
    \end{tabular}
    }
\end{table}

\begin{table}[!p]
    \caption{{\textbf{{\sf \method} boosts quality of downstream decisions for \RTWOGENCNM - evaluated with ChexBert} Results show per-class, macro and micro averaged precision, recall and $F_1$.}}
    \label{tab:results-rcmn-chexbert}
    \centering
    \scalebox{0.95}{
        \begin{tabular}{lrrrrrrrrr}
            \toprule
                & \multicolumn{3}{c}{{\RTWOGENCNM}}
                & \multicolumn{3}{c}{{\RTWOGENCNM (fine-tuned)}}
                & \multicolumn{3}{c}{{\method}}
            \\
          \cmidrule(lr){2-4} \cmidrule(lr){5-7} \cmidrule(lr){8-10}
         \sc {Pathology} & {$Pr$} & {$Rc$} & {$F_1$} & {$Pr$} & {$Rc$} & {$F_1$} & {$Pr$} & {$Rc$} & {$F_1$}
            \\
        \midrule
        {Enlarged Cardiomediastinum} & {12.68} & {\textbf{6.21}} & {\textbf{8.33}} & {8.43} & {4.83} & {6.14} & {8.75} & {4.83} & {6.22} \\
        {Cardiomegaly} & {60.59} & {50.25} & {54.94} & {61.87} & {\textbf{54.35}} & {57.87} & {\textbf{64.03}} & {53.20} & {\textbf{58.12}} \\
        {Lung Opacity} & {46.86} & {\textbf{22.94}} & {30.80} & {46.08} & {15.19} & {22.84} & {\textbf{50.18}} & {22.29} & {\textbf{30.87}} \\
        {Lung Lesion} & {0.00} & {0.00} & {0.00} & {20.00} & {1.02} & {1.94} & {\textbf{66.67}} & {\textbf{2.04}} & {\textbf{3.96}} \\
        {Edema} & {43.33} & {9.06} & {14.99} & {44.64} & {17.42} & {25.06} & {\textbf{54.40}} & {\textbf{23.69}} & {\textbf{33.01}} \\
        {Consolidation} & {18.75} & {6.98} & {10.17} & {20.45} & {10.47} & {13.85} & {\textbf{25.58}} & {\textbf{12.79}} & {\textbf{17.05}} \\
        {Pneumonia} & {\textbf{24.00}} & {8.00} & {\textbf{12.00}} & {18.75} & {\textbf{8.00}} & {11.21} & {13.89} & {6.67} & {9.01} \\
        {Atelectasis} & {28.39} & {30.87} & {29.58} & {\textbf{32.46}} & {30.60} & {\textbf{31.50}} & {30.41} & {\textbf{32.24}} & {31.30} \\
        {Pneumothorax} & {11.11} & {6.67} & {8.33} & {\textbf{20.00}} & {6.67} & {10.00} & {\textbf{21.43}} & {\textbf{10.00}} & {\textbf{13.64}} \\
        {Pleural Effusion} & {\textbf{80.90}} & {31.93} & {45.79} & {72.82} & {33.26} & {45.66} & {74.22} & {\textbf{37.03}} & {\textbf{49.41}} \\
        {Pleural Other} & {0.00} & {0.00} & {0.00} & {0.00} & {0.00} & {0.00} & {0.00} & {0.00} & {0.00} \\
        {Fracture} & {0.00} & {0.00} & {0.00} & {0.00} & {0.00} & {0.00} & {0.00} & {0.00} & {0.00} \\
        {Support Devices} & {77.87} & {53.58} & {63.48} & {\textbf{80.31}} & {47.16} & {59.42} & {80.71} & {\textbf{54.50}} & {\textbf{65.06}} \\
        {No Finding} & {9.51} & {65.96} & {16.62} & {9.75} & {71.28} & {17.16} & {\textbf{11.27}} & {\textbf{77.66}} & {\textbf{19.68}} \\
        \midrule
        {MACRO} & {29.57} & {20.89} & {21.07} & {31.11} & {21.45} & {21.62} & {\textbf{35.82}} & {\textbf{24.07}} & {\textbf{24.09}} \\
        {MICRO} & {42.27} & {31.34} & {36.00} & {41.98} & {30.72} & {35.48} & {\textbf{44.68}} & {\textbf{34.31}} & {\textbf{38.82}} \\
        \bottomrule
    \end{tabular}
    }
\end{table}

\begin{table}[htbp]
    \caption{{\textbf{{\sf \method} boosts quality of downstream decisions for \RTWOGEN - evaluated with ChexBert} Results show per-class, macro and micro averaged precision, recall and $F_1$.}}
    \label{tab:results-rgen-chexbert}
    \centering
    \scalebox{0.95}{
        \begin{tabular}{lrrrrrrrrr}
            \toprule
                & \multicolumn{3}{c}{{\RTWOGEN}}
                & \multicolumn{3}{c}{{\RTWOGEN (fine-tuned)}}
                & \multicolumn{3}{c}{{\method}}
            \\
          \cmidrule(lr){2-4} \cmidrule(lr){5-7} \cmidrule(lr){8-10}
         \sc {Pathology} & {$Pr$} & {$Rc$} & {$F_1$} & {$Pr$} & {$Rc$} & {$F_1$} & {$Pr$} & {$Rc$} & {$F_1$}
            \\
        \midrule
        {Enlarged Cardiomediastinum} & {8.82} & {2.07} & {3.35} & {7.69} & {2.07} & {3.26} & {7.14} & {\textbf{3.45}} & {\textbf{4.65}} \\
        {Cardiomegaly} & {55.52} & {\textbf{31.36}} & {\textbf{40.08}} & {59.91} & {22.33} & {32.54} & {\textbf{63.01}} & {40.56} & {\textbf{49.35}} \\
        {Lung Opacity} & {\textbf{46.67}} & {\textbf{9.05}} & {\textbf{15.16}} & {45.24} & {6.14} & {10.81} & {46.84} & {14.38} & {\textbf{22.00}} \\
        {Lung Lesion} & {0.00} & {0.00} & {0.00} & {0.00} & {0.00} & {0.00} & {0.00} & {0.00} & {0.00} \\
        {Edema} & {47.10} & {22.65} & {30.59} & {\textbf{58.40}} & {\textbf{25.44}} & {\textbf{35.44}} & {49.57} & {19.86} & {28.36} \\
        {Consolidation} & {0.00} & {0.00} & {0.00} & {\textbf{100.00}} & {\textbf{1.16}} & {\textbf{2.30}} & {40.00} & {2.33} & {\textbf{4.40}} \\
        {Pneumonia} & {\textbf{50.00}} & {6.67} & {11.76} & {23.08} & {4.00} & {6.82} & {37.50} & {\textbf{8.00}} & {\textbf{13.19}} \\
        {Atelectasis} & {\textbf{35.86}} & {\textbf{28.42}} & {\textbf{31.71}} & {31.12} & {20.49} & {24.71} & {32.02} & {28.96} & {30.42} \\
        {Pneumothorax} & {0.00} & {0.00} & {0.00} & {16.67} & {6.67} & {9.52} & {\textbf{23.81}} & {\textbf{16.67}} & {\textbf{19.61}} \\
        {Pleural Effusion} & {\textbf{78.67}} & {\textbf{36.81}} & {\textbf{50.15}} & {73.98} & {32.15} & {44.82} & {71.55} & {\textbf{36.81}} & {48.61} \\
        {Pleural Other} & {0.00} & {0.00} & {0.00} & {0.00} & {0.00} & {0.00} & {0.00} & {0.00} & {0.00} \\
        {Fracture} & {0.00} & {0.00} & {0.00} & {0.00} & {0.00} & {0.00} & {0.00} & {0.00} & {0.00} \\
        {Support Devices} & {77.95} & {46.06} & {57.90} & {77.23} & {42.94} & {55.19} & {\textbf{79.94}} & {\textbf{50.46}} & {\textbf{61.87}} \\
        {No Finding} & {9.79} & {\textbf{85.11}} & {\textbf{17.56}} & {8.91} & {\textbf{85.11}} & {16.13} & {\textbf{9.96}} & {78.72} & {\textbf{17.68}} \\
        \midrule
        {MACRO} & {29.31} & {19.16} & {18.45} & {\textbf{35.87}} & {17.75} & {17.25} & {32.95} & {\textbf{21.44}} & {\textbf{21.44}} \\
        {MICRO} & {40.22} & {26.05} & {31.62} & {36.92} & {22.35} & {27.84} & {\textbf{41.97}} & {\textbf{29.19}} & {\textbf{34.43}} \\
        \bottomrule
    \end{tabular}
    }
\end{table}

\end{document}